\documentclass[sigconf,authorversion,nonacm,anonymous=false]{acmart}
\AtBeginDocument{%
  \providecommand\BibTeX{{%
    \normalfont B\kern-0.5em{\scshape i\kern-0.25em b}\kern-0.8em\TeX}}}

\setcopyright{acmcopyright}
\copyrightyear{2024}
\acmYear{2024}
\acmDOI{XXXXXXX.XXXXXXX}

\acmConference[WSDM '2024]{WSDM}{Mar 04--08
  2024}{Yucatán, Mexico}
  
\usepackage{caption}
\usepackage{subcaption}
\usepackage{multirow}

\begin{document}

% \title{Language Models as Knowledge Graphs}
\title{Rethinking Language Models as Symbolic Knowledge Graphs}

%: Is Bigger Better?}

\author{Vishwas Mruthyunjaya, Pouya Pezeshkpour, Estevam Hruschka, Nikita Bhutani}
\email{vishwas, pouya, estevam, nikita@megagon.ai}
% \author{Pouya Pezeshkpour}
% \email{pouya@megagon.ai}
% \author{Nikita Bhutani}
% \email{nikita@megagon.ai}
\affiliation{%
  \institution{Megagon Labs}
  \streetaddress{P.O. Box 1212}
  \city{Mountain View}
  \state{CA}
  \country{USA}
  \postcode{94043}
}


\begin{abstract}

Symbolic knowledge graphs (KGs) play a pivotal role in knowledge-centric applications such as search, question answering and recommendation. As contemporary language models (LMs) trained on extensive textual data have gained prominence, researchers have extensively explored whether the parametric knowledge within these models can match up to that present in knowledge graphs. Various methodologies have indicated that enhancing the size of the model or the volume of training data enhances its capacity to retrieve symbolic knowledge, often with minimal or no human supervision. Despite these advancements, there is a void in comprehensively evaluating whether LMs can encompass the intricate topological and semantic attributes of KGs, attributes crucial for reasoning processes. 

% \estevamc{The following sentence is confusing. Are we going to show LLMs efficacy as KGs and when and how they should be augmented, in this paper? Maybe we can drop it from the abstract.} To ensure the widespread adoption of LMs, it is imperative to grasp their efficacy as KGs and discern when and how they should be augmented. 

In this work, we provide an exhaustive evaluation of language models of varying sizes and capabilities. We construct nine qualitative benchmarks that encompass a spectrum of attributes including symmetry, asymmetry, hierarchy, bidirectionality, compositionality, paths, entity-centricity, bias and ambiguity. Additionally, we propose novel evaluation metrics tailored for each of these attributes. Our extensive evaluation of various LMs shows that while these models exhibit considerable potential in recalling factual information, their ability to capture intricate topological and semantic traits of KGs remains significantly constrained. We note that our proposed evaluation metrics are more reliable in evaluating these abilities than the existing metrics. Lastly, some of our benchmarks challenge the common notion that larger LMs (e.g., GPT-4) universally outshine their smaller counterparts (e.g., BERT).

% Large language models (LLMs) have emerged as powerful tools for tackling a wide range of tasks. This is primarily attributed to the vast knowledge stored in their parameters, enabling them to even construct knowledge graphs with remarkable accuracy. However, to ensure their widespread adoption to new domains and further enhance their performance, it becomes essential to comprehend when and how they should leverage information from knowledge graphs. In this study, we delve into the inherent capabilities of LLMs in understanding both topological and semantical patterns readily available in knowledge graphs. To achieve this, we develop a qualitative benchmark that measures nine different characteristics. Our evaluation of various LLMs on this benchmark reveals that while these models demonstrate great potential, they fall short in accurately grasping any of these patterns. Interestingly, smaller LLMs like BERT outperform their larger counterparts, such as GPT-4, by up to ...\% in capturing some of these characteristics. ...
\end{abstract}

\maketitle

\section{Introduction}

\begin{figure}
  \centering
  \includegraphics[width=\linewidth]{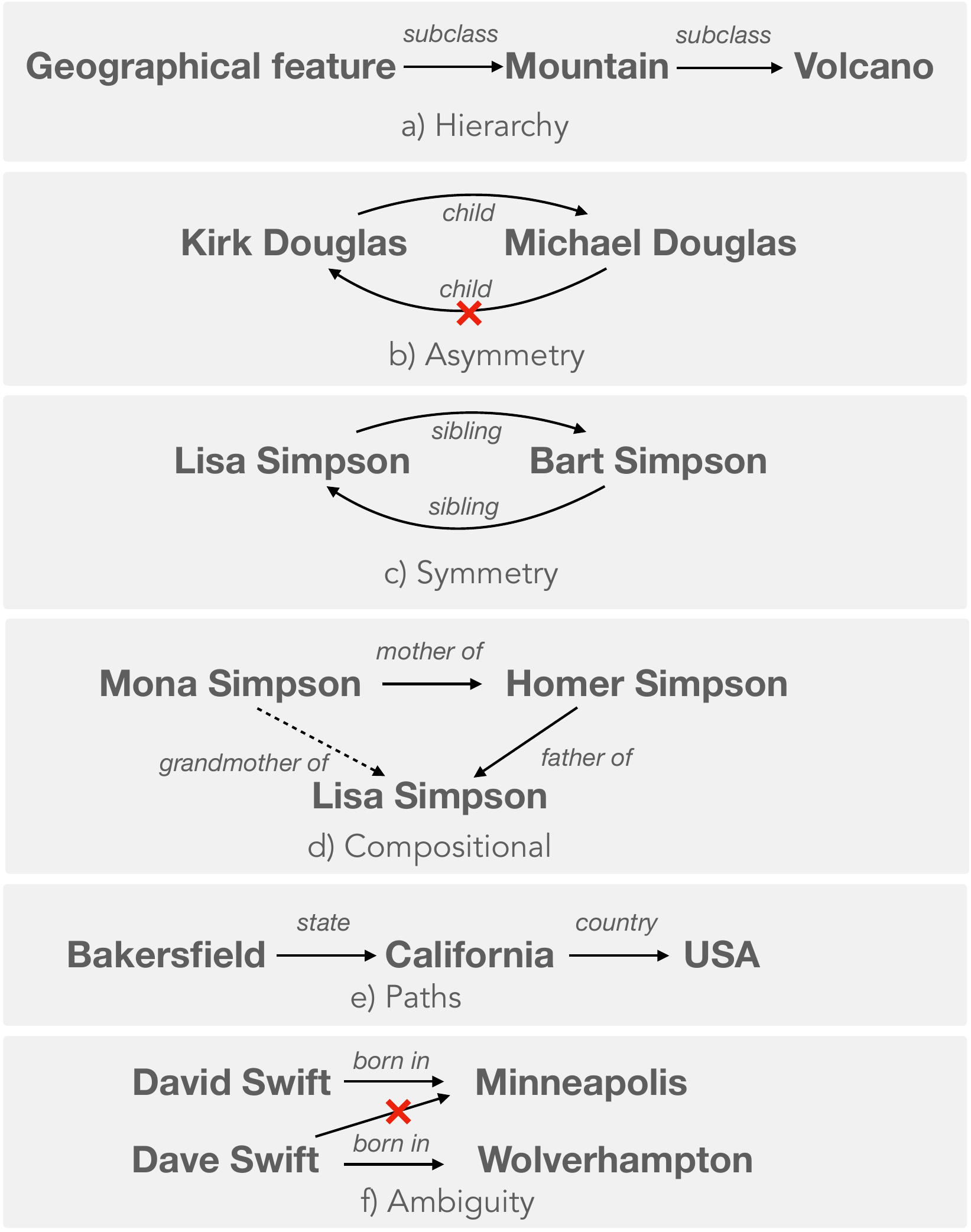}
  \caption{Examples of topological and semantic patterns in real-world KGs.}
  \label{fig:intro_examples}
\end{figure}

Symbolic knowledge graphs (KGs) such as Wikidata ~\cite{vrandevcic2014wikidata}, DBPedia ~\cite{auer2007dbpedia} and Freebase ~\cite{bollacker2008freebase} form the cornerstone of a myriad of applications spanning search engines, question-answering systems and recommendation systems. These applications lean on the structured representation offered by the KGs to access specific pieces of information within them and perform complex reasoning tasks. Recent years have witnessed swift progress in language models (LMs) and their rapidly evolving capabilities. A widely accepted notion is that LMs, pre-trained on extensive textual corpora, hold significant potential to replace symbolic KGs and serve as adaptable repositories of knowledge.

%%%%%%%%%%%%%%%%
\begin{figure*}[t]
  \centering
  \includegraphics[width=0.9\linewidth]{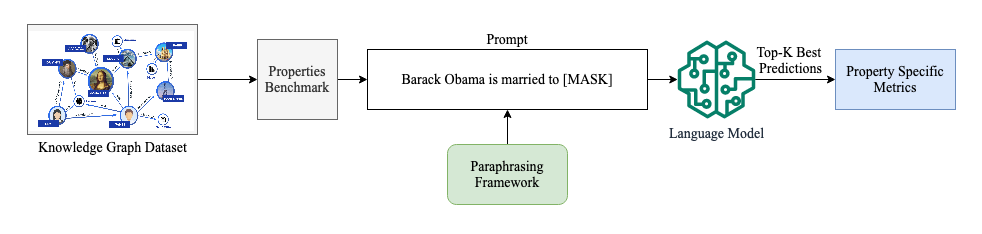}
  \caption{We evaluate LMs capability in capturing KG attributes in each of our benchmarks by measuring the proposed metric for each sample after considering various paraphrases of the facts in the sample ensuring an accurate assessment of LMs knowledge.}
  \label{fig:eval}
\end{figure*}
%%%%%%%%%%%%%%%%%

Stemming from the influential LAMA paper ~\cite{petroni2019language}, a multitude of endeavours have delved into investigating how adeptly do modern LMs encode world knowledge and how to effectively retrieve it. These works unravel the prowess of modern LMs in distilling specific knowledge through various techniques, including prompting in both discrete and continuous forms ~\cite{shin2020autoprompt,li2021prefix,lester2021power} and in-context learning ~\cite{brown2020language,cohen2023crawling}. The extracted knowledge, typically presented as an array of independent triples or a subgraph, is compared with a reference set of triples from a real-world KG. However, we contend that such an evaluation framework falls short in capturing the nuanced attributes of KGs.

KGs have topological and semantic attributes that establish the reliability of the information in them, facilitate easy access and aggregation of information, and enable complex reasoning to be carried out effectively. To achieve parity between LMs and KGs, it is vital to evaluate these attributes. Figure ~\ref{fig:intro_examples} shows some of the topological and semantic patterns in KGs. For example,  taxonomic information such as \textit{hierarchy} is widely adopted for effective completion and retrieval in KGs ~\cite{zhang2020learning,xie2016representation,sarrafzadeh2017improving}. Similarly, semantic constraints such as \textit{symmetry} and \textit{asymmetry} ensure both the reliability of information in the KG and uniformity in the responses to queries over the KG. For example, they can help with providing 
consistent answers for queries involving symmetric relations such as \textit{``Lisa Simpson is sibling of [MASK]''} (answer: Bart Simpson) and \textit{``Bart Simpson is sibling of [MASK]''} (answer: Lisa Simpson). In fact, several knowledge embedding and inference algorithms leverage these semantic constraints. Furthermore, patterns such as \textit{composition} and \textit{paths} form the core of complex reasoning required for question answering over KGs ~\cite{trivedi2017lc,xiong2017deeppath,saxena2020improving,zhu2022neural}. Lastly, the symbolic representation of information in KGs alleviates potential ambiguities arising from lexically-similar entities. To the best of our knowledge, none of the previous evaluation frameworks comprehensively investigate these attributes when establishing the equivalence of LMs and KGs.

Our goal is to thoroughly assess the performance of LMs of varying sizes in capturing the topological and semantic attributes of real-world KGs. To this end, we construct nine novel benchmarks based on the T-REx dataset, a subset of Wikidata triples ~\cite{elsahar2018t}. T-REx triples have been aligned with Wikipedia, a widely employed text corpus for the pre-training of LMs. Specifically, we sample triples from the T-REx dataset for the following attributes: \textit{symmetry}, \textit{asymmetry}, \textit{hierarchy}, \textit{bidirectional}, \textit{compositional}, \textit{paths}, \textit{entity-centric}, \textit{bias}, and \textit{ambiguity}. Each of these benchmarks comprises approximately 1000 examples, forming a comprehensive set for evaluation. Existing metrics (i.e., precision@1) focus primarily on individual triples and are inadequate in providing a dependable assessment for the proposed benchmarks. For example, when evaluating a LM's capability to understand a symmetric relation (e.g. \textit{``Bart Simpson is sibling of Lisa Simpson''}), we would want to discount the model's ability to retrieve  one triplet only (\textit{``Bart Simpson is sibling of Lisa Simpson''} or \textit{``Lisa Simpson is sibling of Bart Simpson''}). To address these limitations, we introduce novel evaluation metrics tailored for each of these distinctive benchmarks.

% para 6: Summary of experimental findings
Our investigation reveals several noteworthy findings. (i) Even the largest of the LMs (GPT-4 \cite{openai2023gpt-4}) achieves only an average of 23.7\% hit@1 on the proposed benchmarks, compared to up to 50\% precision@1 on existing LAMA benchmarks. (ii) LMs may find it relatively simpler to retrieve independent KG triples compared to effectively capturing the nuanced topological and semantic attributes embedded within the KG. (iii) Larger LMs are not universally better than their smaller counterparts. For example, GPT-4 is outperformed by BERT \cite{devlin2018bert} on \textit{bidirectional}, \textit{compositional} and \textit{ambiguity} benchmarks. (iv) Conventional evaluation metrics are only weakly correlated with our proposed metrics. Based on these insights, we conclude  that several significant dimensions of KGs have largely been disregarded during the evaluation of LMs as potential knowledge repositories. Our introduced benchmarks and evaluation metrics hold the potential to guide future research and applications where LMs serve as KGs. We will make our code, benchmarks, and evaluation metrics publicly accessible to foster wider collaboration and exploration.

% bullets -- contributions

\section{Methodology}\label{sec:methodology}

In this section, we introduce our approach to construct benchmarks that capture topological and semantic attributes of KGs and our methodology to test these attributes in language models. 

We use the T-REx dataset ~\cite{elsahar2018t} to create the benchmarks since it provides an extensive collection of Wikipedia abstracts paired with triples. It also has a wide variety of relations that are helpful in producing a diverse set of benchmarks. We sample triples from this dataset to produce benchmarks for \textit{symmetry}, \textit{asymmetry}, \textit{hierarchy}, \textit{bidirectional}, \textit{compositional}, \textit{paths}, 
\textit{entity-centric}, \textit{bias}, and \textit{ambiguity}. Notably, our benchmark design diverges from prior approaches. Instead of individual triples, we devise examples wherein a single benchmark comprises multiple related triples. For instance, both \textit{(Lisa Simpson, is sibling of, Bart Simpson) } and \textit{(Bart Simpson, is sibling of, Lisa Simpson)} form an example within the symmetry benchmark. This unique structure necessitates the development of new evaluation metrics, as traditional metrics tailored to individual triples cannot effectively assess the LMs' understanding of KG attributes.

%%%%%%
\begin{table*}[t]
\small
  \caption{Data statistics of created benchmarks.}
  \label{tab:benchmark}
  \begin{tabular}{ccccccc}
    \toprule
    \bf Benchmark &\bf Num. Triples&\bf Num. Predicates&\bf Num. Entities&\bf Num Triples per Predicate&\bf Example (relation/pattern) \\
    \midrule
     Symmetry&1401&11&1393&127.37&spouse\\
     Asymmetry&8993&90&7870&99.92&date of birth\\
     Bidirectional&6079&116&5387&52.40&instance of -> has part\\
     Hierarchy (2-hop)&8320&1&8320&8320& staple food -> bread\\
     Composite&50488&118&43214&45.16& father -> spouse\\
     Paths (2-hop)&9106&225&8940&40.47& record label -> founder\\
     Entity-Centric&5406&32&4553&168.94&occupation\\
     Ambiguous&750&20&750&37.5&place of birth\\
     Bias&732&4&524&183&sex or gender\\
    \bottomrule
  \end{tabular}
\end{table*}
%%%%%

Figure \ref{fig:eval} shows our evaluation framework\footnote{The knowledge graph image is from \cite{LetTheMachinesLearn}.}. It involves transforming each triplet within an example from a given benchmark into a cloze statement. Subsequently, we employ this statement as a query to prompt a language model for the masked tokens. As an illustration, the triplet \textit{(Lisa Simpson, is sibling of, Bart Simpson) } transforms into \textit{``Lisa Simpson is a sibling of [MASK]''} by masking the object entity. To enhance the robustness of the prompts and reduce prompt sensitivity, we generate variants of the initial prompts using a paraphrase model, Parrot \footnote{\url{https://huggingface.co/prithivida/parrot_paraphraser_on_T5}}, that leverage T5. We further create more variants by randomly replacing tokens in each generated paraphrase with random synonyms. The next step involves comparing the top-k predictions of the language model on the prompts against the true masked entity. The ultimate evaluation metric is then computed based on these comparative scores. We provide details on benchmark creation and evaluation metrics in the next section.

\section{Benchmark and Evaluation}

Given the collection of triples from T-REx dataset, we sample the triples that match various topological and semantic patterns. Each benchmark focuses on a distinct pattern and requires a unique evaluation metric. Table ~\ref{tab:benchmark} shows the statistics of the benchmarks created.

For each benchmark, let $M$ be the target language model to be evaluated. Let $(A, r, B)$ represent a triple in the dataset where $A$ is subject entity, $B$ is the object entity and $r$ is the predicate between them. Let $res$ represent the accuracy of a given triple; this value is 1 if $M$ correctly predicts the object and is 0 otherwise. This metric can be extended to include top-k predictions of the model.

\subsection{Symmetry}

KGs use symmetric constraints to specify that the object entity should also link back to the subject entity. In other words, if the relation $r$ holds between entities $A$ and $B$, then the relation should also hold between $B$ and $A$. Formally,

% Extracting symmetric relations from a triples dataset involves identifying relationships between entities where the relation holds in both directions. If the relation "r" holds between entities "A" and "B", then it is expected that the relation "r" also holds between entities "B" and "A". 

\begin{equation}
    \begin{aligned}
      \forall A, B \in r \rightarrow (A, B)\ and\ r \rightarrow (B, A) \\
      then,\ (A, r, B) \Rightarrow (B, r, A)
    \end{aligned}
\end{equation}

To accurately identify predicates that are symmetric in T-REx, we extract predicates that are associated with at least 50 triples, and at least 50\% of those triples to be symmetric (we found these values extracting the most diverse and accurate set of symmetric predicates). Given the symmetric predicates, we sample at most 200 triples per predicate. 

Let $res_1$ indicate the score for $(A, r, B)$ and $res_2$ indicate the score for $(B, r, A)$. We then compute the symmetric metric for an example as follows: 

% Let A, r, and B be the triples representing the subject, relation, and object, respectively. Let $res_1$ and $res_2$ be the outcomes of evaluating the symmetric relation "r" between A and B in the given order and reversed order, respectively. The metric can be represented in mathematical notation as:
\begin{equation}
    \begin{aligned}
      % \forall A, B \in X(ArB \Leftrightarrow BrA) \\
      % and,\ M:\ Language Model \\
      % then,\ M(A, r, B) \Rightarrow res_1 \\
      % M(B, r, A) \Rightarrow res_2 \\
      % \\
      symmetric\ metric = \frac{res_1 \land res_2}{res_1 \lor res_2}
    \end{aligned}
\end{equation}

In other words, the symmetric metric discounts the model's ability to correctly retrieve at least one of the triples in the example. By doing this, we can effectively evaluate the model's ability to understand symmetric relationships alone. We report the mean symmetric score across all examples in the benchmark.

% The symmetric metric evaluates the symmetry based on the proportion of cases where both $res_1$ and $res_2$ are true, compared to the cases where at least one of res1 or res2 is true. A higher value of the symmetric metric indicates a higher level of symmetry between the evaluated conditions or relationships.
\subsection{Asymmetry}

The inverse of symmetric relations are asymmetric relations wherein a relation $r$ only holds between $A$ and $B$, and not between $B$ and $A$. Formally, 

% An asymmetric relationship is the opposite of a symmetric relationship i.e., the relationship between entities is asymmetric if the relation only holds in one direction. If the relation "r" exists between entities "A" and "B", it is expected that this relation does not hold between "B" and "A". 
\begin{equation}
    \begin{aligned}
      \forall A, B \in r 
      \\
      (A, r, B) \Rightarrow \neg (B, r, A)
    \end{aligned}
\end{equation}

To identify asymmetric predicates, adopting a similar strategy as symmetry predicates, we expect them to be associated with at least 25 triples and at least 50\% of the triples to be asymmetric. Since there are many more asymmetric predicates compared to symmetric ones, this time, we sample at most 100 triples per predicate. 

% Let A, r, and B be the triples representing the subject, relation, and object, respectively. Let $res_1$ and $res_2$ be the outcomes of evaluating the asymmetric relation "r" between A and B in the given order and reversed order, respectively. The metric can be represented in mathematical notation as:

Let $res_1$ indicate the score for $(A, r, B)$ and $res_2$ indicate the score for $(B, r, A)$. We then compute the asymmetric metric for an example as follows: 

\begin{equation}
    \begin{aligned}
      % \forall A, B \in X(ArB) \\
      % and,\ M:\ Language Model \\
      % then,\ M(A, r, B) \Rightarrow res_1 \\
      % M(B, r, A) \Rightarrow res_2 \\
      % \\
      asymmetric\ metric = {res_1 \land \neg(res_2)}
    \end{aligned}
\end{equation}

This ensures the metric reflects the models' understanding of asymmetric relations correctly. We report the mean asymmetric score across all examples in the benchmark.

% The asymmetric metric evaluates the relation symmetry based on the logical AND of res1 being true and res2 being false. It indicates that for the given context, if both res1 is true and res2 is false, then the relation symmetry can be considered asymmetric.

\subsection{Bidirectional/Inverse}

Two predicates $r_1$ and $r_2$ are inverse if they link the same subject and object but in reverse order. In other words, triples $(A, r_1, B)$ and $(B, r_2, A)$ indicate $r_1$ and $r_2$ are inverse to each other. We extract such predicates using the same strategy as symmetric and asymmetric relations.  

% For a relation "r1" that exists between two entities "A" and "B", if there is a unique relation "r2" that exists between "B" and "A", then relations "r1" and "r2" are said to be inverse or entities "A" and "B" have a bidirectional relationship. 
% \begin{equation}
%     \begin{aligned}
%       r1(A, B) \Rightarrow r2(B, A)
%     \end{aligned}
% \end{equation}

To evaluate inverse relations, let $res_1$ and $res_2$ be the accuracy scores for $(A, r_1, B)$ and $(B, r_2, A)$, respectively. We then compute the inverse metric as follows: 

% Let A and B be the entities (subject and/or object), and r1 and r2 be the bidirectional relations representing the entities A and B--$r1(A, B)\ and\ r2(B, A)$. Let $res_1$ and $res_2$ be the outcomes of evaluating the bidirectional relation "r1" between A and B and "r2" between B and A. The metric can be represented in mathematical notation as:
\begin{equation}
    \begin{aligned}
      % \forall r1(A, B) \Rightarrow r2(B, A) \\
      % where,\ r1 \neq r2 \\
      % and,\ M:\ Language Model \\
      % then,\ M(A, r1, B) \Rightarrow res_1 \\
      % M(B, r2, A) \Rightarrow res_2 \\
      % \\
      inverse\ metric = \frac{res_1 \land res_2}{res_1 \lor res_2}
    \end{aligned}
\end{equation}

In other words, the inverse metric captures the model's ability to correctly predict both triples while discounting its ability to retrieve at least one of the triples in the example correctly. We report the mean inverse score across all examples in the benchmark.
% The bidirectional metric evaluates the bi-directionality based on the proportion of cases where both $res_1$ and $res_2$ are true, compared to the cases where at least one of $res_1$ or $res_2$ is true. A higher value of the bidirectional metric indicates a higher level of bi-directionality between the evaluated relationships.

\subsection{Hierarchy}

Hierarchical relations capture an important topological aspect of a KG. In the T-REx dataset, these are represented by \textit{is subclass of} predicate. To evaluate LM's ability to capture hierarchy, we sample 2-hop hierarchical relations from the dataset. Formally, each example in the benchmark is a set of triples $(B, r, A)$ and $(C, r, B)$ where $r$ is \textit{is subclass of}. We randomly sample 1000 such examples from the dataset. 

% Hierarchical relations from a triples dataset represent a hierarchical structure or order amongst the entities. These relations typically indicate a "part-of" or "is-a" relationship, where one entity is a subset or a subclass of another entity. In our specific benchmark creation for hierarchical relation, we have identified "is subclass of" relation from the T-REx dataset. Additionally, the mathematical representation of the hierarchical relation extraction technique is represented as:
% \begin{equation}
%     \begin{aligned}
%     \forall r \in "is\ subclass\ of" \\
%     then,\ X(CrB, BrA) \Rightarrow (CrBrA) \\
%     where,\ A \rightarrow root/parent \\
%     B \rightarrow child\ of\ A \\
%     C \rightarrow child\ of\ B\ and \\
%     \qquad grandchild\ of\ A
%     \end{aligned}
% \end{equation}

To evaluate this benchmark, let $res_1$, $res_2$ and $res_3$ be the accuracy scores for $(B, r, A)$, $(C, r, B)$ and $(C, rr, A)$, respectively where $rr$ captures two levels of being \textit{is subclass of}. We then compute the hierarchical metric as follows: 

% Let A, B, and C be the entities (subject/object) in a triples dataset. For any relationship "r" in the "is subclass of" subset, if there exists $(C, r, B)$ $(B, r, A)$, then the combined relationship $(C, r, B, r, A)$ must also hold true. This can be evaluated using the following hierarchical metric for 2-hop relations:
\begin{equation}
    \begin{aligned}
      %   \forall r \in "is\ subclass\ of" \\
      %   where,\ X(CrB, BrA) \Rightarrow (CrBrA) \\
      %   and,\ M:\ Language Model \\
      %   then,\ M(B, r, A) \Rightarrow res_1 \\
      %   M(C, r, B) \Rightarrow res_2, \\
      %   M(C, r, A) \Rightarrow res_3 \\
      % \\
      hierachical\ metric = \frac{res_1 \land res_2 \land res3}{res_1 \land res_2}
    \end{aligned}
\end{equation}

This metric evaluates whether the model can infer the 2-hop hierarchical relation $(C, rr, A)$ given it can correctly retrieve the 1-hop relations 
$(B, r, A)$ and $(C, r, B)$. We report the mean hierarchical score across all examples in the benchmark.
 
% The hierarchical metric evaluates the hierarchical relationships based on the proportion of cases where $res_1$, $res_2$, and $res_3$ are all true compared to the cases where $res_1$ and $res_2$ are true. A higher value of the hierarchical metric indicates a higher level of hierarchical consistency or pattern in the evaluated conditions or relationships.

\subsection{Composite}

Compositionality in KGs is an important aspect of reasoning tasks in KG completion and querying. A relation $(A, r_3, C)$ is composite if there also exist triples $(A, r_1, B)$ and $(B, r_2, C)$. For example, \textit{father} and \textit{spouse} relations imply \textit{mother} relation. To mine meaningful compositional patterns from the KG, we first identify the top-10 popular categories from the KG based on \textit{instance of} predicate. Politicians, mathematicians, actors, and locations were among the most popular categories. We then collect facts about entities from these categories such that there exists a composite relationship of the form $(A, r_1, B)$, $(B, r_2, C)$, and $(A, r_3, C)$. We further sample from this collection based on the frequency of appearance for each predicate combination. For each predicate combination, we sample 100 compositional facts. If no high-frequent combinations are found, we select a random sample of 1000 compositional facts. 

% To create the composite benchmark, first, we identify popular categories in the knowledge graph (KG) based on instance-of relationships. These categories include politicians, mathematicians, actors, and others. The next step is to select triples about entities within these categories, ensuring the existence of explicit compositional relationships in the form of $(A, r_1, B)$, $(B, r_2, C)$, and $(A, r_3, C)$. Next, we further investigate if there are significant predicate combinations (r1, r2 -> r3). If such combinations exist, then we sample a selection of 100 compositional facts for each combination. However, if no significant predicate combinations are found, a random sample of 1000 compositional facts are created instead from the T-REx dataset.

% We sample 1000 such examples from the T-REx dataset.

% In a composite relation, the presence of one relation is dependent on the presence of another relation. Therefore, for relations r1, r2, and r3 to be composite, we identify if relation "r3" is composed of relations "r1" and "r2".
% \begin{equation}
%     \begin{aligned}
%     \forall A, B, C:\ r1(A, B) \land r2(B, C) \Rightarrow r3(A,C)
%     \end{aligned}
% \end{equation}

To evaluate compositionality, we test the transitive property as follows:

% Let r1, r2, and r3 be unique relations and A, B, and C be unique entities. The specific combination of relations (r1, r2, r3) and entities (A, B, C) such as $r1(A, B),\ r2(B, C),\ and\ r3(A,C)$. We can test the compositional metric of relations (r1, r2, r3) by evaluating their transitive property as follows:
\begin{equation}
    \begin{aligned}
      %   \forall A, B, C:\ r1(A, B) \land r2(B, C) \Rightarrow r3(A,C) \\
      %   where,\ r1 \neq r2 \neq r3 \\
      %   and,\ M:\ Language Model \\
      %   then,\ M(A, r1, B) \Rightarrow res_1 \\
      %   M(B, r2, C) \Rightarrow res_2, \\
      %   M(A, r3, C) \Rightarrow res_3 \\
      % \\
      compositional\ metric = \frac{res_1 \land res_2 \land res3}{res_1 \land res_2}
    \end{aligned}
\end{equation}

\noindent
where $res_1$, $res_2$ and $res_3$ are the accuracy scores for $(A, r_1, B)$, $(B, r_2, C)$ and $(A, r_3, C)$, respectively. We report the mean compositional score across all examples in the benchmark.
% This compositional metric represents the proportion of successful outcomes where all three sub-results, $res_1$, $res_2$, and $res_3$, are achieved (numerator) compared to the proportion of successful outcomes where only the first two sub-results, $res_1$ and $res_2$, are achieved (denominator). This helps to determine the transitive property that confirms the compositionality of relationships.
\subsection{Path}

Path queries are integral to most question-answering benchmarks based on KGs. Exploring how LMs understand paths can provide a deeper understanding of how they learn to internally connect multiple entities. For example, \textit{place of birth} and \textit{country} can help answer questions about the country of birth of a person. 

To construct the benchmark, we follow a similar procedure as the compositional benchmark. Specifically, we first identify popular categories in the KG and select 50 random entities from each of the categories. We then find 2-hop paths from these entities and sample further based on predicate combinations that are frequent. If there were no frequent predicate combinations, we sampled random 10 paths for each entity. Each example in the benchmark is of the form $(A, r_1, B)$ and $(B, r_2, C)$ such that $r_1$ and $r_2$ are not the same. 

To compute the paths metric, let $res_1$ and $res_2$ be the scores for  $(A, r_1, B)$ and $(B, r_2, C)$, respectively. We compute the metric as follows: 

% For our paths benchmark, first, the task was to identify popular categories in KG, including but not limited to politicians, mathematicians, and actors. To accomplish this, we selected 50 entities from each category and proceeded to find 1-hop and 2-hop paths from these entities. For all the 2-hop paths that contained significant predicate combinations, we sampled 100 facts for each predicate combination. If there were no significant predicate combinations, we sampled random 10 paths for each entity.

% Multi-path relationships in a triples dataset are those relationships between entities that can be connected through multiple paths in the dataset. Exploring these relationships can provide a deeper understanding of the connections between entities and uncover hidden patterns and relationships that may not be immediately apparent from a single path. For instance, from equation (6), relation "r1" and "r2" are considered multi-path if there exists entities "A", "B", and "C" where entity "B" is connected to both entities "A" and "C" through relation "r1" and "r2" respectively.
% \begin{equation}
%     \begin{aligned}
%     \{ r1, r2 \} \rightarrow 2-hop\ path\ relation \\
%     if,\ \forall A, B, C:\ r1(A, B) \land r2(B, C)
%     \end{aligned}
% \end{equation}

\begin{equation}
    \begin{aligned}
      % \forall r1(A, B) \Rightarrow r2(B, C) \\
      % where,\ r1 \neq r2 \\
      % and,\ M:\ Language Model \\
      % then,\ M(A, r1, B) \Rightarrow res_1 \\
      % M(B, r2, C) \Rightarrow res_2 \\
      % \\
      paths\ metric = \frac{res_1 \land res_2}{res_1 \lor res_2}
    \end{aligned}
\end{equation}

This evaluates the model's ability to compose the two triples together given it can correctly retrieve at least one of the triples. We report the mean paths score across all examples in the benchmark.

\subsection{Entity-centric}

Tasks such as text generation and summarization require an understanding of a broader set of facts about a given entity. We, therefore, consider building an entity-centric benchmark where each example includes a set of triples centered around a given entity. To create this benchmark, we identify the most well-connected entities in the graph. We then randomly sample from these entities. Furthermore, we sample 20 triples for each entity.  

To compute the entity-centric metric for example, we simply compute the average number of triples in the gold set that could be collectively predicted by the LM. We then report the mean entity-centric score over all examples in the benchmark.
% Entity-centric benchmark data consists of entities that are popular in a triples dataset and are those that occur frequently or are highly connected to other entities in the dataset. To determine popularity, we analyzed the frequency of occurrence and connectivity of subject and object entities in the dataset. We count the number of times an entity appears as a subject or object and calculate the number and strength of its connections with other entities. We then set a threshold based on entity frequencies or connectivity scores. Those that surpass the threshold are considered the most popular entities within the dataset.
% \begin{equation}
%     \begin{aligned}
%     \forall e \in E,\ P = \{ (F(e) > Th) \lor (C(e) > Th) \} \\
%     where,\ P: \text{Popular Entities} \\
%     E: \text{Entities in the dataset} \\
%     F: \text{Frequency} \\
%     C: \text{Connectivity} \\
%     Th: \text{Threshold}
%     \end{aligned}
% \end{equation}

% Let A, B be entities that belong to set of popular entities P. Then, we can evaluate if LMs are capable of capturing information related to relations between popular entities as follows:
% \begin{equation}
%     \begin{aligned}
%       \forall A, B \in P\ where,\ \text{P: Popular Entities}, \\
%       \ r1 \rightarrow (A, B) \\
%       and,\ M:\ Language Model \\
%       then,\ M(A, r1, B) \Rightarrow res_1 \\
%       \\
%       \text{entity-centric metric} = res_1
%     \end{aligned}
% \end{equation}

\subsection{Social Biases}

In order to establish a bias benchmark, we obtained the triples for four specific relations from T-REx: $P21$, which denotes the gender of an individual, $P30$ indicating the location of a particular continent, $P91$ which describes an individual's sexual orientation, and $P140$ which identifies an individual's affiliation with a particular religion. Then, we filtered and normalized the objects in each triple. Subsequently, we only selected triples with subjects that were deemed to be very unpopular, having less than or equal to two links in the T-REx graph \cite{elsahar2018t}. Let us note, that our goal here is to also measure how much LMs will hallucinate if we ask about these sensitive features.
% \begin{equation}
%     \begin{aligned}
%         \forall (s, r, o) \in T-REx \\
%         \text{Filtered Triples} = \forall r \in \{P21, P30, P91, P140\} \\ \cap\ {Triple Subjects} \\
%         \text{Triple Subjects} = s \in \{linkCount(s) \leqslant 2\}
%     \end{aligned}
% \end{equation}

To compute the bias metric, denoting the score for $(A, r, B)$, where $r$ is sensitive relation, by $res$ we introduce the metric as follows: 
\begin{equation}
    \begin{aligned}
      % \forall A, B \in \text{Biased Entities}, \\
      % \ r1 \rightarrow (A, B) \\
      % and,\ M:\ Language Model \\
      % then,\ M(A, r1, B) \Rightarrow res_1 \\
      % \\
      \text{bias metric} = res
    \end{aligned}
\end{equation}
We report the mean bias score across all examples in the benchmark.

\begin{table*}[]
\small
    \caption{Capability of language models in capturing different graph attributes measured by Metric@k (\%).}
    \label{tab: metric percentages}
    \begin{tabular}{ll|ccccccccc}
        \toprule
        &\textbf{Model} & \textbf{Symmetry} & \textbf{Asymmetry} & \textbf{Bidirectional} & \textbf{Hierarchy} & \textbf{Paths} & \textbf{Composite} & \textbf{E-Centric} & \textbf{Bias} & \textbf{Ambiguous} \\ 
        \midrule
        \multirow{5}{*}{\rotatebox[origin=c]{90}{\bf Metric@1}}&BERT & 11.86 & 7.00 & \bf 18.06 & 0.00 & 7.03 & 20.00 & 8.40 & 0.00 & 9.38 \\
        &RoBERTa & 8.51 & 7.50 & 12.50 & 0.00 & 1.83 & 10.00 & 8.80 & 0.00 & \bf 52.63 \\
        &T5 & 8.33 & 2.30 & 0.00 & 0.00 & 0.00 & 0.00 & 1.70 & 0.27 & 0.00 \\
        &GPT-3 & 7.33 & 13.10 & 14.66 & 20.69 &\bf 10.47 & 5.56 & 12.50 & 34.02 & 2.22 \\
        &GPT-4 &\bf 38.19 & \bf 21.80 & 15.34 & \bf 28.57 & 6.37 & \bf 20.45 & \bf 23.10 &\bf 54.10 & 6.02 \\
        \midrule
        \multirow{5}{*}{\rotatebox[origin=c]{90}{\bf Metric@3}}&BERT &\bf 28.23&12.50&17.07&0.00&7.82&42.55&19.50&0.00&30.11\\
        &RoBERTa &17.53&11.50&12.50&25.00&3.19&\bf 52.00&15.20&0.82&\bf 46.15\\
        &T5 &4.00&5.30&1.20&14.29&1.33&0.00&3.70&12.02&0.00\\
        &GPT-3 &21.32&\bf 26.20&\bf 23.25&\bf 49.51&\bf 15.76&48.15&\bf 26.50&\bf 51.78&4.51\\
        &GPT-4 & -&-&-&-&-&-&-&-&-\\
        \midrule
        \multirow{5}{*}{\rotatebox[origin=c]{90}{\bf Metric@5}}&BERT &\bf 32.95 &15.00& 18.75& 0.00&8.25&50.00&24.90&0.00&28.04\\
        &RoBERTa & 21.37&14.60& 10.64&\bf 68.18&4.47&\bf 52.94&20.90&4.51&\bf 38.10\\
        &T5 & 4.26&6.70&7.14&33.33&3.96&0.00& 5.90&28.55&0.00\\
        &GPT-3 & 25.80&\bf 30.80&\bf 24.32&53.79&\bf 21.00&50.88&\bf 36.30&\bf 62.02&6.70 \\
        &GPT-4 & -&-&-&-&-&-&-&-&-\\
        \bottomrule
    \end{tabular}
\end{table*}

\subsection{Ambiguous Entities}

In order to conduct an ambiguity benchmark, a process of collating pairs of entities with similar characteristics (an example of ambiguous pair of entities is provided in Figure \ref{fig:intro_examples}) is necessary. These pairs can consist of either human or non-human entities derived from the fact-checking component of the Amber dataset \cite{chen2021evaluating}. In order to extract the respective relations and linked objects per entity, their individual Wikipedia pages are utilized. The resulting data is then filtered to only include shared relations that are also present in the LAMA relations dataset \cite{petroni2019language}. Each example in the benchmark is of the form $(A, r, B)$ and $(\overline{A}, r, C)$ where A represents the well-known entity and $\overline{A}$ corresponds to the lesser-known entity. This methodology ensures a systematic and comprehensive approach to evaluating ambiguity. 
% Mathematical representation of ambiguous entities are as follows:.
% \begin{equation}
%     \begin{aligned}
%     \forall A \neq \overline{A} \in r1 \\
%     \{A, \overline{A} \} \rightarrow ambiguous \\
%     if,\ A, B:\ r1(A, B) \\
%     \overline{A}, C: r1(\overline{A}, C)
%     \end{aligned}
% \end{equation}
% Let r1 be the relation that consist of two similar subjects A and $\overline{A}$ whose objects are B and C, respectively. Then, 
To calculate LMs capability in capturing the right information between these ambiguous entity pairs, we use the following metric:
\begin{equation}
    \begin{aligned}
    % \forall A \neq \overline{A}, but\ A \sim \overline{A} \\
    %   where,\ (A, B)\ and\ (\overline{A}, C) \in r1 \\
    %   and,\ M:\ Language Model \\
    %   then,\ M(A, r1, B) \Rightarrow res_1 \\
    %   M(\overline{A}, r1, C) \Rightarrow res_2 \\
    %   \\
      ambiguous\ metric = \frac{res_1 \land res_2}{res_1 \lor res_2}
    \end{aligned}
\end{equation}
where $res_1$, $res_2$ are the accuracy scores for $(A, r, B)$ and $(\overline{A}, r, C)$, respectively. We report the mean ambiguous score across all examples in the benchmark.

\section{Experiments}

We evaluate an array of robust language models of varying sizes against our benchmarks. This includes smaller models such as BERT (bert-base-uncased), RoBERTa (roberta-base), T5 (t5-base) and more advanced larger models such as GPT-3 (text-davinci-003), and GPT-4. Next, we describe our experimental set-up, followed by a deeper analysis of the performance of various language models. 

% In our experiments to evaluate language models performance on our benchmarks, we employ a range of powerful language models including . Furthermore, we leverage a T5 LM-based paraphrasing framework called Parrot to generate diverse prompts. These prompts are then used as inputs to the language models for benchmark evaluation. The following sections will give a brief look into the model setup and prompt generation.

\subsection{Setup}

\noindent{\textbf{Prompting}} We rely on initial prompts for each relation in the benchmarks. We use the prompts for the T-REx dataset from \citet{petroni2019language}. Since this only covers a subset of the relations in our benchmark, we manually define templates for any new relations. For example, relation \textit{spouse} is associated with a prompt \textit{``<ENT0> is spouse of <ENT1>''}. For a target triple, we replace  \textit{<ENT0>} with the subject of the triple and \textit{<ENT1>} with the respective $MASK$ tag of the LM. We then paraphrase these prompts using the strategy described in Section~\ref{sec:methodology}. For each triple, we rely on the list of paraphrases to probe an LM. We then consider the best prediction for a given list of prompts as the final prediction to calculate the metric. We observe that since GPT-3 and GPT-4 are much more robust to wordings of the prompt, 10 randomly sampled paraphrased prompts were adequate for their accurate assessment. In contrast, we used all the paraphrased prompts for BERT, RoBERTa, and T5.

%% -- Nikita -- condensing this a bit since it is duplicate from Sec 2. 
% \textit{Initial Prompt:} for every triple, (subject, relation, object), in the benchmark, we extract initial prompts from T-REx respective to the relation in a triple. For example, if the relation is \textit{spouse}, then, the prompt extracted from T-REx would be of the form \textit{"<ENT0> is spouse of <ENT1>"}. Then, we replace $\textit{<ENT0>}$ with \textit{subject} and \textit{<ENT1>} with models respective $MASK$ tags. 

% \textit{Paraphrasing Framework:} to extend and produce a variety of prompts from the initial prompt extracted from the T-REx dataset, we utilize the Parrot library that leverages T5 for paraphrasing. Subsequently, we expand the paraphrases by replacing tokens in each generated variant with random synonyms, excluding the subject tokens to maintain coherence. 

% To clarify, the list of paraphrases and the initial prompt will be used as input for language models. The label for the masked token will be the object in the triple and only the best prediction for a given list of paraphrases (prompts) is considered for the metric calculation. The goal here is to get an accurate assessment of LMs knowledge despite their sensitivity to the wording of a prompt.  
% We observe that since GPT-3 and GPT-4 are much more robust to minor changes in the prompt, 10 paraphrase samples were adequate for their accurate assessment, whereas BERT, RoBERTa, and T5 required to receive the entire list (less than 100 paraphrase samples).

\noindent{\textbf{BERT}} To evaluate BERT \cite{devlin2018bert}, we use the $fill\_mask$ method, a utility provided by Hugging Face\footnote{https://huggingface.co/tasks/fill-mask}. The method initiates by segmenting the input sentence into tokens, identifying the $[MASK]$ token, and then translating these tokens into corresponding ID tensors. The model then predicts the token to replace $[MASK]$, considering the context provided by the surrounding tokens. This method returns a list of $top-k$ (where k=5 in our experiments) predicted tokens along with their associated probabilities. 

Note that since BERT predicts one token per masked token, we handle multi-token entities by replacing them with an equivalent number of $[MASK]$ tokens. For example, for a triple \textit{(Michelle Obama, spouse, Barack Obama)} we initialize the prompt  \textit{``<ENTO> is married to <ENT1>"} as \textit{``Michelle Obama is married to [MASK] [MASK]''}. We then use the average of the accuracy of the multiple tokens as the final score of the prediction.

% It performs masked language modeling where a $[MASK]$ token is replaced in a sentence and the model predicts the original token. The method first tokenizes the input sentence, identifies the $[MASK]$ token, and encodes the tokens into ID tensors. The BERT model then predicts the token to replace the $[MASK]$, considering the context provided by the surrounding tokens. The method returns a list of the $top-k$ (where k=5 in our experiments) predicted tokens along with their probabilities. 

%%%%%%%%%%%%%%%%
\begin{figure*}[]
  \centering
  \includegraphics[width=0.8\linewidth]{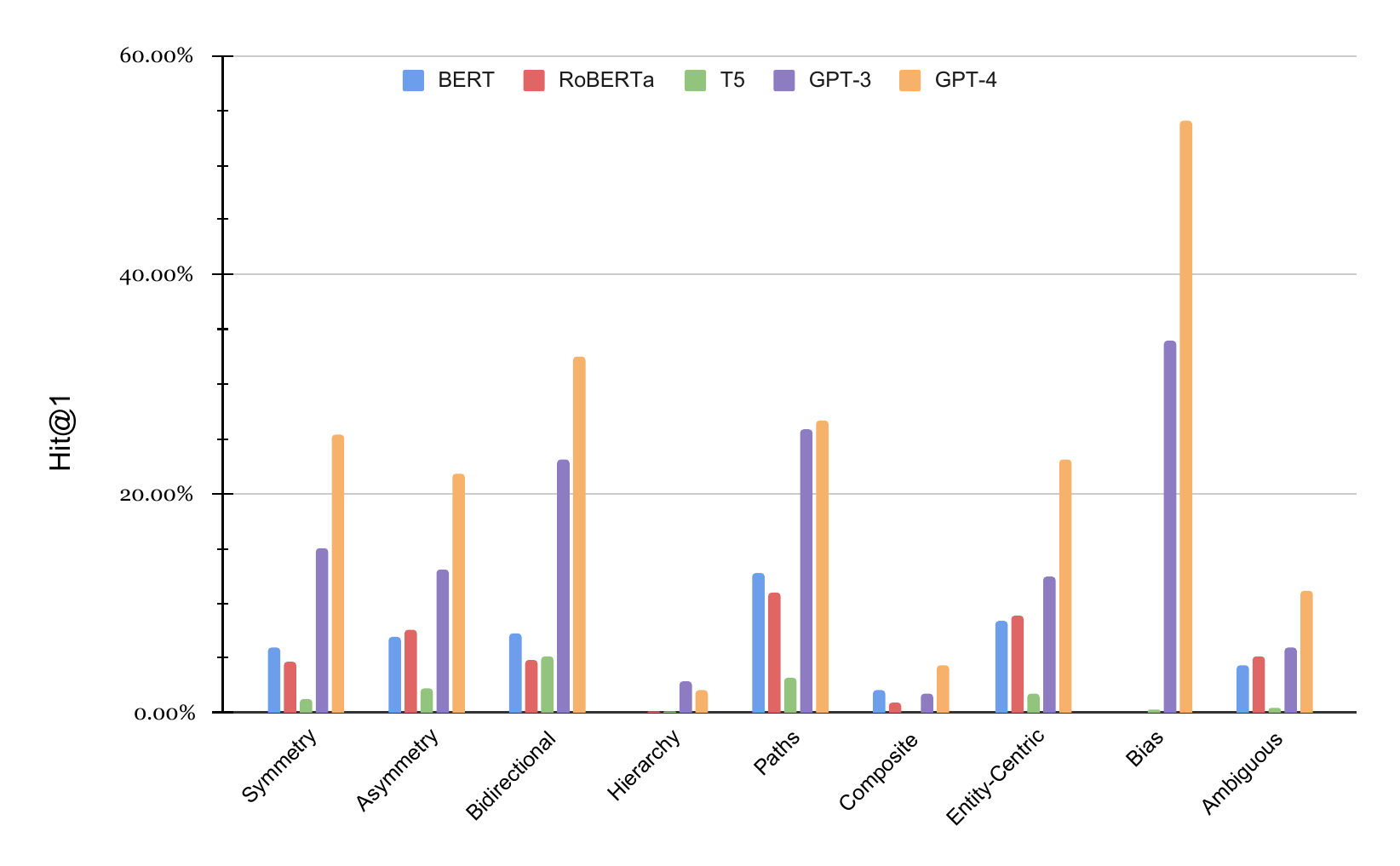}
  \caption{Language models partial score in each benchmark calculated at Hits@1.}
  \label{fig:partial}
\end{figure*}
%%%%%%%%%%%%%%%%%
\begin{figure*}[]
    \centering
        \captionsetup[subfigure]{justification=centering}
        \begin{subfigure}{.3\linewidth}
            \centering
            \includegraphics[width=\textwidth]{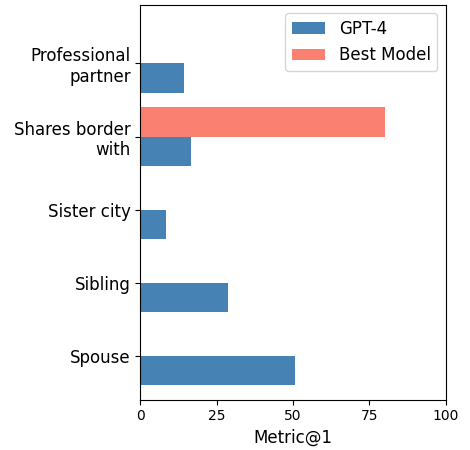}
            \caption{Symmetry.}
            \label{fig:sym}
        \end{subfigure}
        \begin{subfigure}{.3\linewidth}
            \centering
            \includegraphics[width=\textwidth]{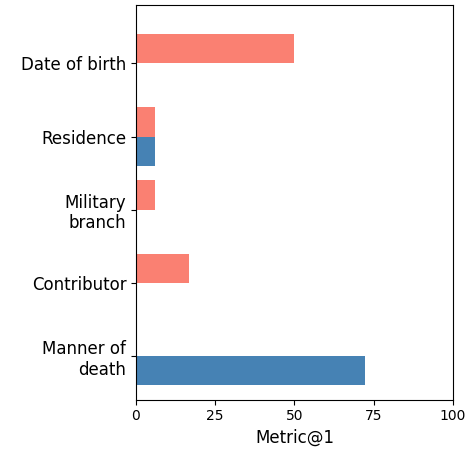}
            \caption{Asymmetry.}
            \label{fig:asym}
        \end{subfigure}
        \begin{subfigure}{.3\linewidth}
            \centering
            \includegraphics[width=\textwidth]{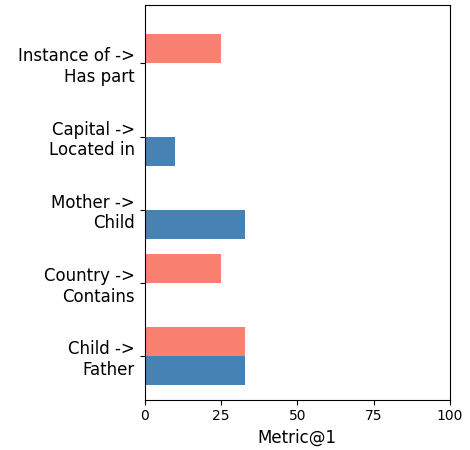}
            \caption{Bidirectional.}
            \label{fig:bid}
        \end{subfigure}
        \begin{subfigure}{.3\linewidth}
            \centering
            \includegraphics[width=\textwidth]{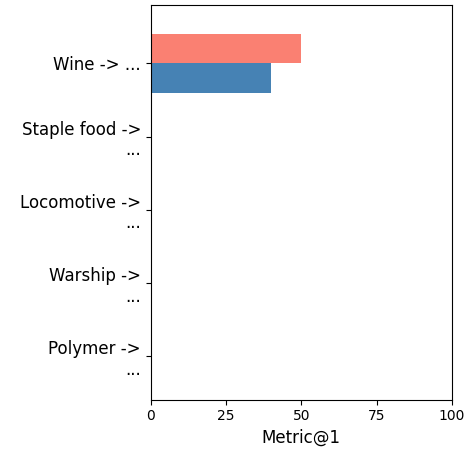}
            \caption{Hierarchy.}
            \label{fig:hier}
        \end{subfigure}
        \begin{subfigure}{.3\linewidth}
            \centering
            \includegraphics[width=\textwidth]{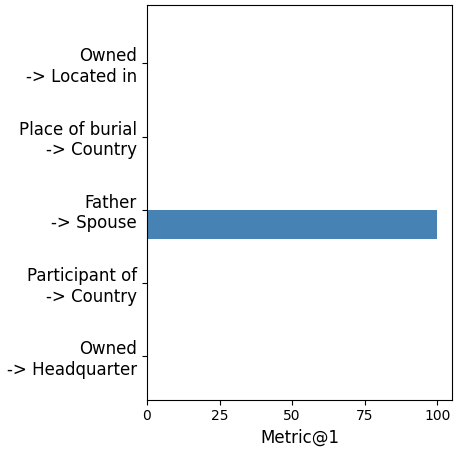}
            \caption{Composite.}
            \label{fig:comp}
        \end{subfigure}
        \begin{subfigure}{.3\linewidth}
            \centering
            \includegraphics[width=\textwidth]{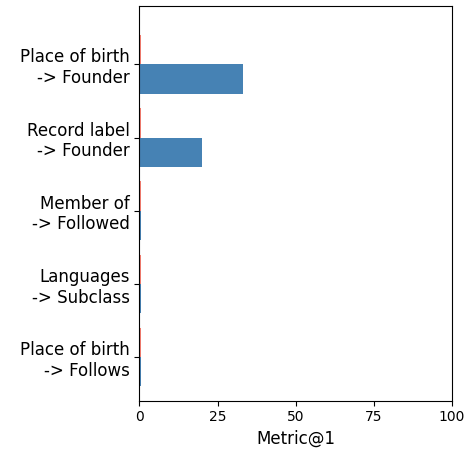}
            \caption{Path.}
            \label{fig:mpath}
        \end{subfigure}
        \begin{subfigure}{.3\linewidth}
            \centering
            \includegraphics[width=\textwidth]{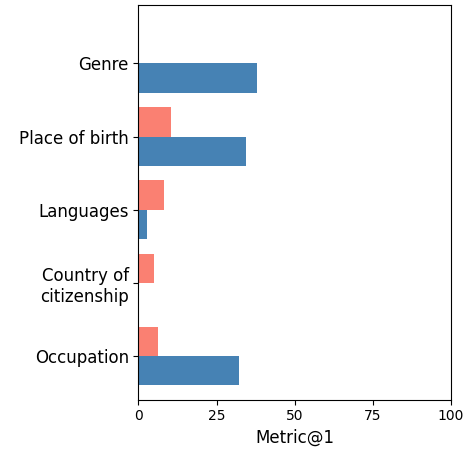}
            \caption{Entity-centric.}
            \label{fig:e-centric}
        \end{subfigure}
        \begin{subfigure}{.3\linewidth}
            \centering
            \includegraphics[width=\textwidth]{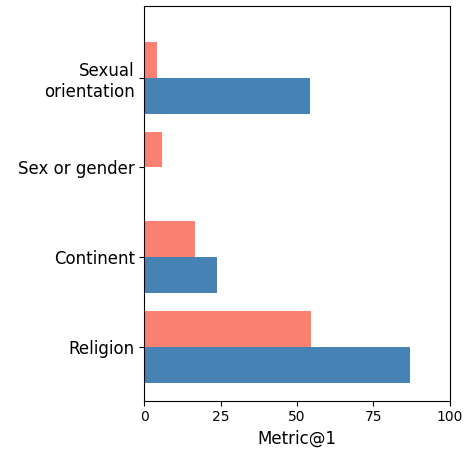}
            \caption{Bias.}
            \label{fig:bias}
        \end{subfigure}
        \begin{subfigure}{.3\linewidth}
            \centering
            \includegraphics[width=\textwidth]{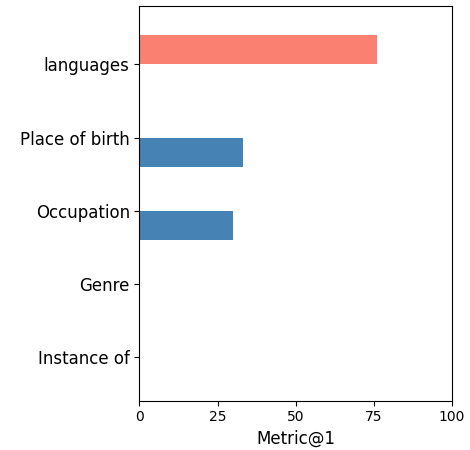}
            \caption{Ambiguous.}
            \label{fig:ambig}
        \end{subfigure}
        \caption{\textbf{Per-relation/pattern breakdown} of language models performance for each benchmark. A missing bar means Metric@1=0. Also, For (c) Bidirectional, GPT-4 was the best model for the pattern Mother -> Child.}
    \label{fig:per-relation}
\end{figure*}
%%%%%%%%%%

%%%%%%%%%%%%%%%%
\begin{figure*}[t]
  \centering
  \includegraphics[width=0.9\linewidth]{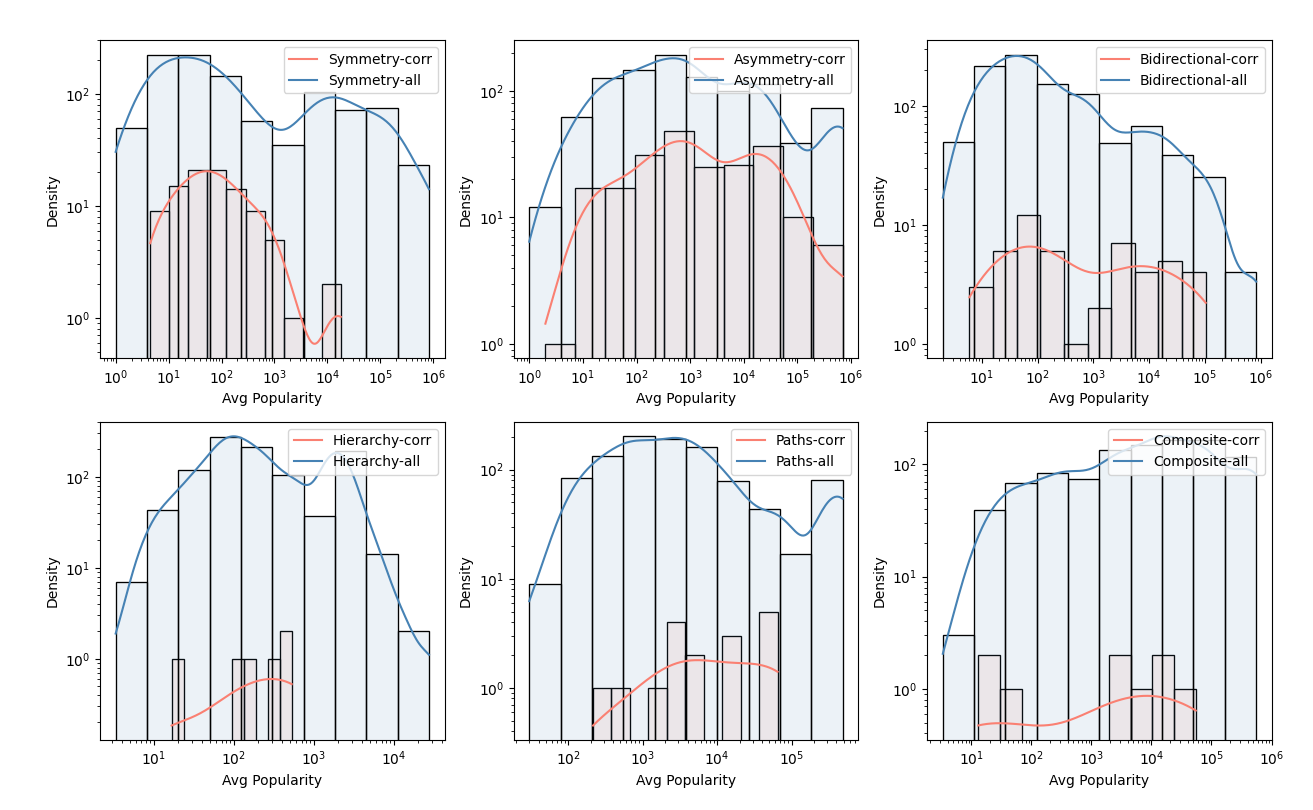}
  \caption{Histogram plot over average popularity of entities in each example across different benchmarks. We plot two distributions for \textit{all} examples and only correctly (\textit{corr}) predicted examples in each benchmark.}
  \label{fig:hist}
\end{figure*}
%%%%%%%%%%%%%%%%%

% Note that the masked token for BERT is represented as \textit{"[MASK]"}. And, for multi-token labels, the masked token will represent the number of words in the label. For instance, if $(sub, rel, obj)$ = \textit{(Michelle Obama, spouse, Barack Obama)} and the prompt is \textit{"Michelle Obama is married to <ENT1>."} before the mask token is annotated. Then, after mask-token annotation, the prompt would appear as follows: \textit{Michelle Obama is married to [MASK] [MASK] .} 

\noindent{\textbf{RoBERTa}} We follow similar approach to evaluate RoBERTa \cite{liu2019roberta} with the exception of using \textit{<mask>} as the masked token.  

% \textit{RoBERTa:} similar to BERT, RoBERTa \cite{liu2019roberta} uses the $fill\_mask$ method from Hugging Face. However, the biggest distinction is that the masked token is represented as \textit{"<mask>"}. The functionality of $fill\_mask$ remains the same---encoding tokens to ID tensors, multi-token masking, predicting the masked token, and predicting top 5 objects.

\noindent{\textbf{T5}} Unlike BERT and RoBERTa, T5 \cite{raffel2020exploring} is designed primarily of text-to-text generation tasks. We rely on the \textit{fill\_mask} utility for probing T5 with one distinction. We use the special \textit{<extra\_id\_0>} token as placeholder for the object entity in the prompt. We then tokenize the input prompt, pass the encoded inputs to T5 and generate top-k predictions. Note that since T5 can generate spans for the mask, it can easily support multi-token entities.

% \textit{T5:} performing masked token prediction using the Hugging Face T5 model \cite{raffel2020exploring} involves using the same approach as the \textit{fill\_mask} method. However, it is important to note that T5 is primarily designed for text-to-text tasks rather than masked language modeling (MLM) like BERT or RoBERTa.
% To perform masked token prediction with the T5 model, we first replace the target token with the special \textit{"<extra\_id\_0>"} token. This token is used as a placeholder for masked tokens in T5. Then, after tokenizing the input sentence, we pass the encoded inputs to the T5 model, to generate top-k predictions for the masked token.

\noindent{\textbf{GPT-3}} We use the OpenAI API with \textit{text-davinci-003} model \cite{ouyang2022training} for text completion. For a fair comparison to other models, we use  
$[MASK]$ for the target entities in the prompt. We then extend the prompt to include a task description for GPT-3. An example prompt to GPT-3 looks like \textit{``Replace [MASK] with the most probable words. Michelle Obama is married to [MASK]."}. We additionally use \textit{logprobs} as the input parameter to retrieve top-5 predictions for the generated text. We then use these scores to compare the generated text with the ground-truth entities.

% \textit{GPT-3:} to generate text completion for a prompt using the OpenAI GPT-3 language model, we utilised the \textit{text-davinci-003} model \cite{ouyang2022training}. To simulate HuggingFace's \textit{fill\_mask} method, we modify the prompt by replacing the "<ENT1>" tag with "[MASK]" in a sentence like "Michelle Obama is married to <ENT1>". Before passing the prompt to the GPT-3 model, we further add a description to the prompt such as \textit{"Replace [MASK] with the most probable words. Michelle Obama is married to [MASK]."} Upon incorporating \textit{logprobs} input parameter into the model, \textit{text-davinci-003} gives us only the top 5 best matches from the generated text. This enables us to compare the generated text with the label (object) to evaluate the GPT-3 model predictions.

\noindent{\textbf{GPT-4}} Similar to GPT-3, we use the OpenAI API to generate chat-based text completions with GPT-4 \cite{openai2023gpt-4}. We extend the prompt to include task instructions as in GPT-3. To ensure the accuracy of the GPT-4 assessment, we experimented with several different task instructions, such as \textit{``List the top 20 best possible entities to fill the [MASK]''}. Regrettably, the predictions exhibited a certain insensitivity to these different prompts. Consequently, we decided to maintain uniformity by employing the same prompt descriptions as in the GPT-3 context. Additionally, we found that setting the temperature parameter to 0 improved the accuracy of predictions, albeit at the cost of dampening the diversity of the top-k predictions. We, therefore, only report hits@1 scores for GPT-4. 

% \textit{GPT-4:} We also adopt GPT-4 \cite{openai2023gpt-4} to generate chat-based text completions through the OpenAI API. While there are some differences in how GPT-4 operates compared to GPT-3, the majority of the functional aspects remain the same. To simulate the \textit{fill\_mask method}, we modify the prompt for the GPT-4 model similarly to how we did for GPT-3. We also experimented with different prompts, such as listing the top 20 best possible entities to fill the [MASK], but found that the results were the same as our original prompting method. Furthermore, upon setting GPT-4 temperature to 0 for accurate generation, the output becomes extremely repetitive for any value of $n$. As a result, to ensure fairness, we only compare GPT-4 in hits@1 with other language models.

\subsection{Main Results}

We present the performances of different LMs on our devised benchmarks, as assessed using the novel metrics we introduced. These results are summarized in Table \ref{tab: metric percentages}.

\noindent{\textbf{Challenging benchmarks}} As can be seen, even the most powerful model GPT-4 achieves an average of 23.77\% (6.02\% - 54\% ) metric@1 score across different benchmarks. In contrast, the best GPT models report up to 64\% precision on the widely used LAMA benchmark \cite{liu2021gpt}. This indicates that our proposed benchmarks are challenging and show a wide gap in the knowledge modeling/retrieval capabilities of large language models. We need models that capture the topological and semantic patterns we proposed as well as effective techniques to recover these patterns from language models.

\noindent{\textbf{Effect of size of language models}} There is a common notion that larger LMs (e.g., GPT-4) are universally more powerful than smaller counterparts (e.g, BERT). While this notion holds true for numerous of our benchmarks, including attributes like symmetry, asymmetry, hierarchy, compositional, entity-centric, and bias, our investigations reveal intriguing insights. We find that larger models face more challenges than their smaller counterparts in specific benchmarks, namely bidirectional relationships, complex paths, and scenarios involving ambiguity.

For the bidirectional benchmark, only BERT demonstrates superior performance compared to GPT-4. While, in the paths benchmark, both BERT and GPT-3 outperform GPT-4. We examine the per-relation breakdown in the later sections to understand this behavior. It is not very surprising that there is a wide performance gap between larger models (GPT-4 and GPT-3) in comparison to smaller models (BERT and RoBERTa). Larger models are known to hallucinate \cite{ji2023survey,du2023improving}. Especially under uncertain and ambiguous scenarios, they tend to confidently generate non-factual information. 

On the other hand, we observe that larger LMs exhibit marked improvements over conventional ones, especially in their aptitude to address and manage bias. This progress is evident from the noticeable disparities observed in the bias benchmark. We conjecture that this is due to the adoption of strategic alignment techniques \cite{ouyang2022training} aimed to alleviate social biases inherent in content produced by large LMs. This aligns with previous discoveries that LMs trained to advocate fairness demonstrate heightened resilience to social biases \cite{ganguli2022red}. Finally, GPT-4 also stands out in the entity-centric benchmark, supporting the overarching assertion that the breadth of knowledge in LMs is closely related to their scale.

\noindent{\textbf{Ranking distribution}} We observe analogous trends in metric@3 and metric@5 when comparing GPT-3 with other LMs, with a few noteworthy deviations. Intriguingly, GPT-3 is consistently surpassed by BERT on the symmetry benchmark. One might anticipate GPT-3, having undergone training on a considerably extensive dataset, to have a stronger grasp of comprehending the symmetry attribute. Conversely, GPT-3 outshines BERT in the bidirectional benchmark, particularly evident in cases with higher values of k. This implies that while GPT-3 may possess a better grasp of bi-directionality compared to BERT, it still does not fully internalize this attribute. A parallel trend is echoed by RoBERTa, whose performance shows a notable enhancement with increasing k values. This is particularly pronounced in the hierarchy and composite benchmarks, where RoBERTa showcases substantial gains of 68.18\% and 42.94\% respectively, ultimately surpassing even GPT-3. This could be attributed to RoBERTa's stronger grasp of these attributes but its lack of factual knowledge, resulting in low metric@1.

% For metric@3 and metric@5, excluding GPT-4, we mostly observe similar pattern when comparing GPT-3 with other LMs.
% In terms of the symmetry attribute, GPT-3 is consistently surpassed by BERT, exhibiting a clear advantage for the latter. This is specifically surprising, as one might expect that GPT-3, trained on a significantly larger volume of data, would have had greater chances to grasp the concept of symmetry due to the increased prevalence of such relationships within the data it has been exposed to. 
% Conversely, within the bidirectional attribute, GPT-3 manages to reclaim the leading position.
% Turning attention to the hierarchy and composite benchmarks, a notable transformation is observed for RoBERTa. As the top-k value is elevated to 5, RoBERTa's performance experiences a substantial leap from 0\% to 68.18\% in the hierarchy attribute, and from 10\% to 52.94\% in the composite attribute outperforming GPT-3. This could be attributed to RoBERTa's stronger grasp of these attributes but its lacking factual knowledge which results in having a very low metric@1 while performing considerably better when increasing the top-k values.

In summary, our evaluation highlights differences in how models internalize different KG attributes. All language models consistently demonstrate inadequate performance across various attributes, implying inherent limitations in capturing KG attributes. Moreover, contradictory to popular belief scales of LMs are not entirely correlated with their ability to capture these attributes.

% In summary, the analysis highlights divergences in how the models address distinct relationship characteristics. Certain models display elevated percentages across multiple categories, underscoring their proficiency in capturing particular types of relationships. Nonetheless, it's important to note that all language models demonstrate consistently inadequate performance across various attributes. This collectively underscores their inherent limitations in effectively capturing attributes within knowledge graphs. Moreover, in contradiction to a common belief, it becomes evident that the scales of language models are not entirely correlated with their ability to capture these attributes.

\subsection{Analysis}

\noindent{\textbf{Effectiveness of proposed metrics}} We hypothesized that simply evaluating triples in isolation is ineffective in measuring the model's capabilities. To support this hypothesis, we consider the partial scores (i.e., denominator) of the proposed metric. These partial scores are simply a conjunction of scores of individual triples in an example in a benchmark. We report these scores in Figure \ref{fig:partial}. As can be seen, these metric shows that GPT-4 significantly outperforms all other models. This is inconsistent with the observations made using our proposed metric. In fact, we find that there is only a very weak correlation between these partial scores and our proposed metrics. We thus conclude that existing metrics that evaluate triples in isolation overestimate the model performance in capturing nuanced KG attributes.

% In order to gain deeper insights into the reasons behind the inability of language models to effectively capture attributes within knowledge graphs, we present the partial scores for each metric. These partial scores represent the denominator in our proposed fractional metrics and the entire metric otherwise, as illustrated in Figure \ref{fig:partial}.
% The data reveals that GPT-4 significantly outperforms other LMs, primarily attributed to its extensive repository of knowledge. Importantly, it's worth noting that there exists a weak correlation between the partial score and our proposed metric. This observation underscores the limitations of existing evaluation metrics in accurately gauging LMs' grasp of KG attributes. Furthermore, our observations consistently indicate that T5 performs less effectively than other language models, highlighting its unsuitability for the task of masked language modeling.

\noindent{\textbf{Relation-wise breakdown}} To better understand the capabilities of LMs we look at the relation-wise performances of LMs across benchmarks. Especially, for each benchmark, we compare the performances (metric@1 score) of the powerful GPT-4 model with the best-performing model. We show the results for the 5 most frequent relations/patterns in Figure \ref{fig:per-relation}.
% We also delve into the performance of language models concerning different relations/patterns within each benchmark. The metric@1 score for the five most frequent relations/patterns is represented in Figure \ref{fig:per-relation}.
This analysis aims to address two key questions: (i) Is the inability of LMs to capture specific attributes uniformly distributed across various relations/patterns, or is it skewed toward certain ones? (ii) Does the smaller model that outperforms GPT-4 exhibit distinct behavioral patterns? 

We find that even when we drill down into specific relations/ patterns, GPT-4 tends to be on par with or outperforms smaller models across most patterns, barring a few exceptions. However, it appears that LM's grasp of KG attributes is not consistent across different relations/patterns. While it can understand an attribute well for specific relations and patterns, it completely misses those attributes in other instances. These variations might originate from insufficient representation of these patterns in textual data. Such inconsistencies can severely affect downstream applications that expect a uniform understanding of topological and semantic patterns, regardless of specific relations/patterns. 

In summary, relation-wise breakdowns not only shed light on the shortcomings of LMs, but also provide an actionable strategy: to improve the LMs performance we can rely on knowledge graphs when concerned with missing relations/patterns for each attribute.

\noindent{\textbf{Performance versus popularity}} Although LMs perform poorly on almost all of our proposed benchmarks, one can argue that even this limited level of achievement is not indicative of a deep understanding of the topological and semantic attributes. Rather, it can be attributed to the fact that LMs have simply memorized the triples in the examples during training.
% Rather, it can be attributed to their training on the triples associated with correctly predicted instances. 
To further investigate the capability of LMs in grasping these proposed attributes, we provide the histogram of the average popularity of entities appearing in each example (calculated using the T-REx knowledge graph) in Figure \ref{fig:hist}. Our analysis is centered exclusively on GPT-4 due to its substantially larger repository of facts in comparison to other models. Additionally, we only consider benchmarks that do not utilize popularity in filtering triples when curating the data. In the histograms, we illustrate the distribution for \textit{all} examples within each benchmark, as well as exclusively for examples where GPT-4's predictions were \textit{correct}, indicating cases where GPT-4's predictions for all triples in an example were correct. 
The histograms reveal a noteworthy observation: the distribution of entity popularity in correctly predicted examples closely mirrors the distribution observed across all examples within each benchmark. This suggests that GPT-4's predictive outcomes are not solely reliant on its factual knowledge but also stem from an understanding of each semantic attribute. 
Furthermore, it appears that correctly predicted examples tend to exhibit a certain minimum level of popularity among entities. However, not only the predictive influence of entity popularity diminish once a specific threshold is reached, but also it seems GPT-4 has a hard time correctly predicting examples with a very high level of average popularity. 

\section{Related Work}

Numerous studies have looked into the equivalence and alignment between language models and knowledge graphs. Some notable works \cite{petroni2019language, safavi2021relational, roberts-etal-2020-much} have demonstrated the capability of language models to internalize vast world knowledge within their parameters. This knowledge can subsequently be retrieved with minimal or no human guidance. Pioneered by \citet{petroni2019language}, most approaches adopt cloze-style prompting, employing hand-crafted discrete prompts \cite{brown2020language, heinzerling-inui-2021-language} or automated prompts \cite{qin-eisner-2021-learning, zhong-etal-2021-factual} to extract and evaluate factual knowledge in LMs. However, these works typically evaluate LMs using independent triples drawn from a real-world KG. Such a simplified setting is inadequate to assess whether LMs can truly replace symbolic KGs on knowledge-intensive tasks. 

Conversely, other studies have embarked on exploring how graphs can be harvested more broadly from LMs \cite{alivanistos2022prompting,hao-etal-2023-bertnet,cohen2023crawling}. The basic idea involves starting with a seed entity and employing a template for each relation to query the LMs and generate subgraphs centered around the seed entity. Further advancements \cite{cohen2023crawling} delve into minimizing the manual effort in the knowledge extraction process by eliminating the need to pre-specify relations of interest. Alternatively, some researchers have evaluated LMs directly on benchmarks designed for knowledge graph completion \cite{veseli2023evaluating, nadkarni2021scientific}. 

Distinguishing itself from prior research, our study presents a unique perspective by offering a comprehensive insight into the internal representation of knowledge within LMs. Instead of evaluating isolated triples or specific tasks, we focus on investigating different topological and semantic attributes of knowledge from symbolic KGs.  Our endeavor is to ascertain whether the internal representation of LMs aptly captures these attributes. We firmly believe that the benchmarks and revelations from our study can serve as guiding principles for future model development, steering them towards embodying these attributes or suggesting avenues to enhance LMs with these attributes.

\section{Conclusion}

Gaining an in-depth comprehension of modern LMs and their extensive capabilities holds paramount importance, given their wide-ranging applications. Specifically, it is imperative to understand their limitations in how they learn, represent and store world knowledge. This can inform techniques to improve the models and discern when and how they should be augmented with external knowledge. Despite the substantial body of work dedicated to assessing the potential substitution of KGs with LMs, it is evident that these investigations tend to overlook the multifaceted aspects of symbolic representation that undermine the significance of KGs across diverse applications. We develop new benchmarks and evaluation metrics to address various topological and semantic attributes of KGs. Our experiments reveal that LMs are still far from fully capturing the topological and semantic attributes of symbolic representation. In fact, for some of the benchmarks smaller LMs outperform the more popular larger LMs. We confer that conducting thorough evaluations of LMs, as exemplified in this paper, is essential for their meaningful progression.

\bibliographystyle{ACM-Reference-Format}
\bibliography{main}


\begin{thebibliography}{37}

%%% the Web address.

\ifx \showCODEN    \undefined \def \showCODEN     #1{\unskip}     \fi
\ifx \showDOI      \undefined \def \showDOI       #1{#1}\fi
\ifx \showISBNx    \undefined \def \showISBNx     #1{\unskip}     \fi
\ifx \showISBNxiii \undefined \def \showISBNxiii  #1{\unskip}     \fi
\ifx \showISSN     \undefined \def \showISSN      #1{\unskip}     \fi
\ifx \showLCCN     \undefined \def \showLCCN      #1{\unskip}     \fi
\ifx \shownote     \undefined \def \shownote      #1{#1}          \fi
\ifx \showarticletitle \undefined \def \showarticletitle #1{#1}   \fi
\ifx \showURL      \undefined \def \showURL       {\relax}        \fi
% invisible to TeX
\providecommand\bibfield[2]{#2}
\providecommand\bibinfo[2]{#2}
\providecommand\natexlab[1]{#1}
\providecommand\showeprint[2][]{arXiv:#2}

\bibitem[Alivanistos et~al\mbox{.}(2022)]%
        {alivanistos2022prompting}
\bibfield{author}{\bibinfo{person}{Dimitrios Alivanistos},
  \bibinfo{person}{Selene~B{\'a}ez Santamar{\'\i}a}, \bibinfo{person}{Michael
  Cochez}, \bibinfo{person}{Jan-Christoph Kalo}, \bibinfo{person}{Emile van
  Krieken}, {and} \bibinfo{person}{Thiviyan Thanapalasingam}.}
  \bibinfo{year}{2022}\natexlab{}.
\newblock \showarticletitle{Prompting as probing: Using language models for
  knowledge base construction}.
\newblock \bibinfo{journal}{\emph{arXiv preprint arXiv:2208.11057}}
  (\bibinfo{year}{2022}).
\newblock


\bibitem[Auer et~al\mbox{.}(2007)]%
        {auer2007dbpedia}
\bibfield{author}{\bibinfo{person}{S. Auer}, \bibinfo{person}{Christian Bizer},
  \bibinfo{person}{Georgi Kobilarov}, \bibinfo{person}{Jens Lehmann},
  \bibinfo{person}{Richard Cyganiak}, {and} \bibinfo{person}{Zachary~G. Ives}.}
  \bibinfo{year}{2007}\natexlab{}.
\newblock \showarticletitle{DBpedia: A Nucleus for a Web of Open Data}. In
  \bibinfo{booktitle}{\emph{Proc. ISWC/ASWC 2007}}.
\newblock
\urldef\tempurl%
\url{https://api.semanticscholar.org/CorpusID:7278297}
\showURL{%
\tempurl}


\bibitem[Bollacker et~al\mbox{.}(2008)]%
        {bollacker2008freebase}
\bibfield{author}{\bibinfo{person}{Kurt Bollacker}, \bibinfo{person}{Colin
  Evans}, \bibinfo{person}{Praveen Paritosh}, \bibinfo{person}{Tim Sturge},
  {and} \bibinfo{person}{Jamie Taylor}.} \bibinfo{year}{2008}\natexlab{}.
\newblock \showarticletitle{Freebase: a collaboratively created graph database
  for structuring human knowledge}. In \bibinfo{booktitle}{\emph{Proc. SIGMOD
  2008}}. \bibinfo{pages}{1247--1250}.
\newblock


\bibitem[Brown et~al\mbox{.}(2020)]%
        {brown2020language}
\bibfield{author}{\bibinfo{person}{Tom Brown}, \bibinfo{person}{Benjamin Mann},
  \bibinfo{person}{Nick Ryder}, \bibinfo{person}{Melanie Subbiah},
  \bibinfo{person}{Jared~D Kaplan}, \bibinfo{person}{Prafulla Dhariwal},
  \bibinfo{person}{Arvind Neelakantan}, \bibinfo{person}{Pranav Shyam},
  \bibinfo{person}{Girish Sastry}, \bibinfo{person}{Amanda Askell},
  {et~al\mbox{.}}} \bibinfo{year}{2020}\natexlab{}.
\newblock \showarticletitle{Language models are few-shot learners}.
\newblock \bibinfo{journal}{\emph{Advances in neural information processing
  systems}}  \bibinfo{volume}{33} (\bibinfo{year}{2020}),
  \bibinfo{pages}{1877--1901}.
\newblock


\bibitem[Chen et~al\mbox{.}(2021)]%
        {chen2021evaluating}
\bibfield{author}{\bibinfo{person}{Anthony Chen}, \bibinfo{person}{Pallavi
  Gudipati}, \bibinfo{person}{Shayne Longpre}, \bibinfo{person}{Xiao Ling},
  {and} \bibinfo{person}{Sameer Singh}.} \bibinfo{year}{2021}\natexlab{}.
\newblock \showarticletitle{Evaluating Entity Disambiguation and the Role of
  Popularity in Retrieval-Based NLP}. In \bibinfo{booktitle}{\emph{Proc.
  ACL-IJCNLP 2021}}. \bibinfo{pages}{4472--4485}.
\newblock


\bibitem[Cohen et~al\mbox{.}(2023)]%
        {cohen2023crawling}
\bibfield{author}{\bibinfo{person}{Roi Cohen}, \bibinfo{person}{Mor Geva},
  \bibinfo{person}{Jonathan Berant}, {and} \bibinfo{person}{Amir Globerson}.}
  \bibinfo{year}{2023}\natexlab{}.
\newblock \showarticletitle{Crawling The Internal Knowledge-Base of Language
  Models}. In \bibinfo{booktitle}{\emph{Findings of the Association for
  Computational Linguistics: EACL 2023}}. \bibinfo{pages}{1811--1824}.
\newblock


\bibitem[Devlin et~al\mbox{.}(2018)]%
        {devlin2018bert}
\bibfield{author}{\bibinfo{person}{Jacob Devlin}, \bibinfo{person}{Ming-Wei
  Chang}, \bibinfo{person}{Kenton Lee}, {and} \bibinfo{person}{Kristina
  Toutanova}.} \bibinfo{year}{2018}\natexlab{}.
\newblock \showarticletitle{Bert: Pre-training of deep bidirectional
  transformers for language understanding}.
\newblock \bibinfo{journal}{\emph{arXiv preprint arXiv:1810.04805}}
  (\bibinfo{year}{2018}).
\newblock


\bibitem[Du et~al\mbox{.}(2023)]%
        {du2023improving}
\bibfield{author}{\bibinfo{person}{Yilun Du}, \bibinfo{person}{Shuang Li},
  \bibinfo{person}{Antonio Torralba}, \bibinfo{person}{Joshua~B Tenenbaum},
  {and} \bibinfo{person}{Igor Mordatch}.} \bibinfo{year}{2023}\natexlab{}.
\newblock \showarticletitle{Improving Factuality and Reasoning in Language
  Models through Multiagent Debate}.
\newblock \bibinfo{journal}{\emph{arXiv preprint arXiv:2305.14325}}
  (\bibinfo{year}{2023}).
\newblock


\bibitem[Elsahar et~al\mbox{.}(2018)]%
        {elsahar2018t}
\bibfield{author}{\bibinfo{person}{Hady Elsahar}, \bibinfo{person}{Pavlos
  Vougiouklis}, \bibinfo{person}{Arslen Remaci}, \bibinfo{person}{Christophe
  Gravier}, \bibinfo{person}{Jonathon Hare}, \bibinfo{person}{Frederique
  Laforest}, {and} \bibinfo{person}{Elena Simperl}.}
  \bibinfo{year}{2018}\natexlab{}.
\newblock \showarticletitle{T-rex: A large scale alignment of natural language
  with knowledge base triples}. In \bibinfo{booktitle}{\emph{Proc. LREC 2018}}.
\newblock


\bibitem[Ganguli et~al\mbox{.}(2022)]%
        {ganguli2022red}
\bibfield{author}{\bibinfo{person}{Deep Ganguli}, \bibinfo{person}{Liane
  Lovitt}, \bibinfo{person}{Jackson Kernion}, \bibinfo{person}{Amanda Askell},
  \bibinfo{person}{Yuntao Bai}, \bibinfo{person}{Saurav Kadavath},
  \bibinfo{person}{Ben Mann}, \bibinfo{person}{Ethan Perez},
  \bibinfo{person}{Nicholas Schiefer}, \bibinfo{person}{Kamal Ndousse},
  {et~al\mbox{.}}} \bibinfo{year}{2022}\natexlab{}.
\newblock \showarticletitle{Red teaming language models to reduce harms:
  Methods, scaling behaviors, and lessons learned}.
\newblock \bibinfo{journal}{\emph{arXiv preprint arXiv:2209.07858}}
  (\bibinfo{year}{2022}).
\newblock


\bibitem[Hao et~al\mbox{.}(2023)]%
        {hao-etal-2023-bertnet}
\bibfield{author}{\bibinfo{person}{Shibo Hao}, \bibinfo{person}{Bowen Tan},
  \bibinfo{person}{Kaiwen Tang}, \bibinfo{person}{Bin Ni},
  \bibinfo{person}{Xiyan Shao}, \bibinfo{person}{Hengzhe Zhang},
  \bibinfo{person}{Eric Xing}, {and} \bibinfo{person}{Zhiting Hu}.}
  \bibinfo{year}{2023}\natexlab{}.
\newblock \showarticletitle{{B}ert{N}et: Harvesting Knowledge Graphs with
  Arbitrary Relations from Pretrained Language Models}. In
  \bibinfo{booktitle}{\emph{Findings of ACL 2023}}.
  \bibinfo{publisher}{Association for Computational Linguistics},
  \bibinfo{address}{Toronto, Canada}, \bibinfo{pages}{5000--5015}.
\newblock
\urldef\tempurl%
\url{https://doi.org/10.18653/v1/2023.findings-acl.309}
\showDOI{\tempurl}


\bibitem[Heinzerling and Inui(2021)]%
        {heinzerling-inui-2021-language}
\bibfield{author}{\bibinfo{person}{Benjamin Heinzerling} {and}
  \bibinfo{person}{Kentaro Inui}.} \bibinfo{year}{2021}\natexlab{}.
\newblock \showarticletitle{Language Models as Knowledge Bases: On Entity
  Representations, Storage Capacity, and Paraphrased Queries}. In
  \bibinfo{booktitle}{\emph{Proc. EACL 2021}}. \bibinfo{publisher}{Association
  for Computational Linguistics}, \bibinfo{address}{Online},
  \bibinfo{pages}{1772--1791}.
\newblock
\urldef\tempurl%
\url{https://doi.org/10.18653/v1/2021.eacl-main.153}
\showDOI{\tempurl}


\bibitem[Ji et~al\mbox{.}(2023)]%
        {ji2023survey}
\bibfield{author}{\bibinfo{person}{Ziwei Ji}, \bibinfo{person}{Nayeon Lee},
  \bibinfo{person}{Rita Frieske}, \bibinfo{person}{Tiezheng Yu},
  \bibinfo{person}{Dan Su}, \bibinfo{person}{Yan Xu}, \bibinfo{person}{Etsuko
  Ishii}, \bibinfo{person}{Ye~Jin Bang}, \bibinfo{person}{Andrea Madotto},
  {and} \bibinfo{person}{Pascale Fung}.} \bibinfo{year}{2023}\natexlab{}.
\newblock \showarticletitle{Survey of hallucination in natural language
  generation}.
\newblock \bibinfo{journal}{\emph{Comput. Surveys}} \bibinfo{volume}{55},
  \bibinfo{number}{12} (\bibinfo{year}{2023}), \bibinfo{pages}{1--38}.
\newblock


\bibitem[Lester et~al\mbox{.}(2021)]%
        {lester2021power}
\bibfield{author}{\bibinfo{person}{Brian Lester}, \bibinfo{person}{Rami
  Al-Rfou}, {and} \bibinfo{person}{Noah Constant}.}
  \bibinfo{year}{2021}\natexlab{}.
\newblock \showarticletitle{The Power of Scale for Parameter-Efficient Prompt
  Tuning}. In \bibinfo{booktitle}{\emph{Proc. EMNLP 2021}}.
  \bibinfo{pages}{3045--3059}.
\newblock


\bibitem[Li and Liang(2021)]%
        {li2021prefix}
\bibfield{author}{\bibinfo{person}{Xiang~Lisa Li} {and} \bibinfo{person}{Percy
  Liang}.} \bibinfo{year}{2021}\natexlab{}.
\newblock \showarticletitle{Prefix-Tuning: Optimizing Continuous Prompts for
  Generation}. In \bibinfo{booktitle}{\emph{Proc. ACL-IJCNLP 2021}}.
  \bibinfo{pages}{4582--4597}.
\newblock


\bibitem[Liu et~al\mbox{.}(2021)]%
        {liu2021gpt}
\bibfield{author}{\bibinfo{person}{Xiao Liu}, \bibinfo{person}{Yanan Zheng},
  \bibinfo{person}{Zhengxiao Du}, \bibinfo{person}{Ming Ding},
  \bibinfo{person}{Yujie Qian}, \bibinfo{person}{Zhilin Yang}, {and}
  \bibinfo{person}{Jie Tang}.} \bibinfo{year}{2021}\natexlab{}.
\newblock \showarticletitle{GPT understands, too}.
\newblock \bibinfo{journal}{\emph{arXiv preprint arXiv:2103.10385}}
  (\bibinfo{year}{2021}).
\newblock


\bibitem[Liu et~al\mbox{.}(2019)]%
        {liu2019roberta}
\bibfield{author}{\bibinfo{person}{Yinhan Liu}, \bibinfo{person}{Myle Ott},
  \bibinfo{person}{Naman Goyal}, \bibinfo{person}{Jingfei Du},
  \bibinfo{person}{Mandar Joshi}, \bibinfo{person}{Danqi Chen},
  \bibinfo{person}{Omer Levy}, \bibinfo{person}{Mike Lewis},
  \bibinfo{person}{Luke Zettlemoyer}, {and} \bibinfo{person}{Veselin
  Stoyanov}.} \bibinfo{year}{2019}\natexlab{}.
\newblock \showarticletitle{Roberta: A robustly optimized bert pretraining
  approach}.
\newblock \bibinfo{journal}{\emph{arXiv preprint arXiv:1907.11692}}
  (\bibinfo{year}{2019}).
\newblock


\bibitem[Nadkarni et~al\mbox{.}(2021)]%
        {nadkarni2021scientific}
\bibfield{author}{\bibinfo{person}{Rahul Nadkarni}, \bibinfo{person}{David
  Wadden}, \bibinfo{person}{Iz Beltagy}, \bibinfo{person}{Noah~A Smith},
  \bibinfo{person}{Hannaneh Hajishirzi}, {and} \bibinfo{person}{Tom Hope}.}
  \bibinfo{year}{2021}\natexlab{}.
\newblock \showarticletitle{Scientific language models for biomedical knowledge
  base completion: an empirical study}.
\newblock \bibinfo{journal}{\emph{arXiv preprint arXiv:2106.09700}}
  (\bibinfo{year}{2021}).
\newblock


\bibitem[OpenAI(2023)]%
        {openai2023gpt-4}
\bibfield{author}{\bibinfo{person}{OpenAI}.} \bibinfo{year}{2023}\natexlab{}.
\newblock \showarticletitle{GPT-4 Technical Report}.
\newblock \bibinfo{journal}{\emph{arXiv preprint arXiv:2303.08774}}
  (\bibinfo{year}{2023}).
\newblock


\bibitem[Ouyang et~al\mbox{.}(2022)]%
        {ouyang2022training}
\bibfield{author}{\bibinfo{person}{Long Ouyang}, \bibinfo{person}{Jeffrey Wu},
  \bibinfo{person}{Xu Jiang}, \bibinfo{person}{Diogo Almeida},
  \bibinfo{person}{Carroll Wainwright}, \bibinfo{person}{Pamela Mishkin},
  \bibinfo{person}{Chong Zhang}, \bibinfo{person}{Sandhini Agarwal},
  \bibinfo{person}{Katarina Slama}, \bibinfo{person}{Alex Ray},
  {et~al\mbox{.}}} \bibinfo{year}{2022}\natexlab{}.
\newblock \showarticletitle{Training language models to follow instructions
  with human feedback}.
\newblock \bibinfo{journal}{\emph{Advances in Neural Information Processing
  Systems}}  \bibinfo{volume}{35} (\bibinfo{year}{2022}),
  \bibinfo{pages}{27730--27744}.
\newblock


\bibitem[Petroni et~al\mbox{.}(2019)]%
        {petroni2019language}
\bibfield{author}{\bibinfo{person}{Fabio Petroni}, \bibinfo{person}{Tim
  Rockt{\"a}schel}, \bibinfo{person}{Sebastian Riedel},
  \bibinfo{person}{Patrick Lewis}, \bibinfo{person}{Anton Bakhtin},
  \bibinfo{person}{Yuxiang Wu}, {and} \bibinfo{person}{Alexander Miller}.}
  \bibinfo{year}{2019}\natexlab{}.
\newblock \showarticletitle{Language Models as Knowledge Bases?}
\newblock  (\bibinfo{year}{2019}), \bibinfo{pages}{2463--2473}.
\newblock


\bibitem[Qin and Eisner(2021)]%
        {qin-eisner-2021-learning}
\bibfield{author}{\bibinfo{person}{Guanghui Qin} {and} \bibinfo{person}{Jason
  Eisner}.} \bibinfo{year}{2021}\natexlab{}.
\newblock \showarticletitle{Learning How to Ask: Querying {LM}s with Mixtures
  of Soft Prompts}. In \bibinfo{booktitle}{\emph{Proc. NAACL 2021}}.
  \bibinfo{publisher}{Association for Computational Linguistics},
  \bibinfo{address}{Online}, \bibinfo{pages}{5203--5212}.
\newblock
\urldef\tempurl%
\url{https://doi.org/10.18653/v1/2021.naacl-main.410}
\showDOI{\tempurl}


\bibitem[Raffel et~al\mbox{.}(2020)]%
        {raffel2020exploring}
\bibfield{author}{\bibinfo{person}{Colin Raffel}, \bibinfo{person}{Noam
  Shazeer}, \bibinfo{person}{Adam Roberts}, \bibinfo{person}{Katherine Lee},
  \bibinfo{person}{Sharan Narang}, \bibinfo{person}{Michael Matena},
  \bibinfo{person}{Yanqi Zhou}, \bibinfo{person}{Wei Li}, {and}
  \bibinfo{person}{Peter~J Liu}.} \bibinfo{year}{2020}\natexlab{}.
\newblock \showarticletitle{Exploring the limits of transfer learning with a
  unified text-to-text transformer}.
\newblock \bibinfo{journal}{\emph{The Journal of Machine Learning Research}}
  \bibinfo{volume}{21}, \bibinfo{number}{1} (\bibinfo{year}{2020}),
  \bibinfo{pages}{5485--5551}.
\newblock


\bibitem[Roberts et~al\mbox{.}(2020)]%
        {roberts-etal-2020-much}
\bibfield{author}{\bibinfo{person}{Adam Roberts}, \bibinfo{person}{Colin
  Raffel}, {and} \bibinfo{person}{Noam Shazeer}.}
  \bibinfo{year}{2020}\natexlab{}.
\newblock \showarticletitle{How Much Knowledge Can You Pack Into the Parameters
  of a Language Model?}. In \bibinfo{booktitle}{\emph{Proc. EMNLP 2020}}.
  \bibinfo{publisher}{Association for Computational Linguistics},
  \bibinfo{address}{Online}, \bibinfo{pages}{5418--5426}.
\newblock
\urldef\tempurl%
\url{https://doi.org/10.18653/v1/2020.emnlp-main.437}
\showDOI{\tempurl}


\bibitem[Safavi and Koutra(2021)]%
        {safavi2021relational}
\bibfield{author}{\bibinfo{person}{Tara Safavi} {and} \bibinfo{person}{Danai
  Koutra}.} \bibinfo{year}{2021}\natexlab{}.
\newblock \showarticletitle{Relational world knowledge representation in
  contextual language models: A review}.
\newblock \bibinfo{journal}{\emph{arXiv preprint arXiv:2104.05837}}
  (\bibinfo{year}{2021}).
\newblock


\bibitem[Sarrafzadeh and Lank(2017)]%
        {sarrafzadeh2017improving}
\bibfield{author}{\bibinfo{person}{Bahareh Sarrafzadeh} {and}
  \bibinfo{person}{Edward Lank}.} \bibinfo{year}{2017}\natexlab{}.
\newblock \showarticletitle{Improving Exploratory Search Experience through
  Hierarchical Knowledge Graphs}.
\newblock \bibinfo{journal}{\emph{Proc. ACM SIGIR 2017}}
  (\bibinfo{year}{2017}).
\newblock
\urldef\tempurl%
\url{https://api.semanticscholar.org/CorpusID:24547416}
\showURL{%
\tempurl}


\bibitem[Saxena et~al\mbox{.}(2020)]%
        {saxena2020improving}
\bibfield{author}{\bibinfo{person}{Apoorv Saxena}, \bibinfo{person}{Aditay
  Tripathi}, {and} \bibinfo{person}{Partha Talukdar}.}
  \bibinfo{year}{2020}\natexlab{}.
\newblock \showarticletitle{Improving multi-hop question answering over
  knowledge graphs using knowledge base embeddings}. In
  \bibinfo{booktitle}{\emph{Proc. ACL 2020}}. \bibinfo{pages}{4498--4507}.
\newblock


\bibitem[Seth(2019)]%
        {LetTheMachinesLearn}
\bibfield{author}{\bibinfo{person}{Yashu Seth}.}
  \bibinfo{year}{2019}\natexlab{}.
\newblock \bibinfo{booktitle}{\emph{Introduction to question answering over
  knowledge graphs}}.
\newblock
\urldef\tempurl%
\url{https://yashuseth.wordpress.com/2019/10/08/introduction-question-answering-knowledge-graphs-kgqa/}
\showURL{%
Retrieved August 10, 2023 from \tempurl}


\bibitem[Shin et~al\mbox{.}(2020)]%
        {shin2020autoprompt}
\bibfield{author}{\bibinfo{person}{Taylor Shin}, \bibinfo{person}{Yasaman
  Razeghi}, \bibinfo{person}{Robert~L Logan~IV}, \bibinfo{person}{Eric
  Wallace}, {and} \bibinfo{person}{Sameer Singh}.}
  \bibinfo{year}{2020}\natexlab{}.
\newblock \showarticletitle{AutoPrompt: Eliciting Knowledge from Language
  Models with Automatically Generated Prompts}. In
  \bibinfo{booktitle}{\emph{Proc. EMNLP 2020}}. \bibinfo{pages}{4222--4235}.
\newblock


\bibitem[Trivedi et~al\mbox{.}(2017)]%
        {trivedi2017lc}
\bibfield{author}{\bibinfo{person}{Priyansh Trivedi}, \bibinfo{person}{Gaurav
  Maheshwari}, \bibinfo{person}{Mohnish Dubey}, {and} \bibinfo{person}{Jens
  Lehmann}.} \bibinfo{year}{2017}\natexlab{}.
\newblock \showarticletitle{Lc-quad: A corpus for complex question answering
  over knowledge graphs}. In \bibinfo{booktitle}{\emph{Proc. ISWC 2017}}.
  Springer, \bibinfo{pages}{210--218}.
\newblock


\bibitem[Veseli et~al\mbox{.}(2023)]%
        {veseli2023evaluating}
\bibfield{author}{\bibinfo{person}{Blerta Veseli}, \bibinfo{person}{Sneha
  Singhania}, \bibinfo{person}{Simon Razniewski}, {and}
  \bibinfo{person}{Gerhard Weikum}.} \bibinfo{year}{2023}\natexlab{}.
\newblock \showarticletitle{Evaluating Language Models for Knowledge Base
  Completion}. In \bibinfo{booktitle}{\emph{Proc. ESWC 2023}}. Springer,
  \bibinfo{pages}{227--243}.
\newblock


\bibitem[Vrande{\v{c}}i{\'c} and Kr{\"o}tzsch(2014)]%
        {vrandevcic2014wikidata}
\bibfield{author}{\bibinfo{person}{Denny Vrande{\v{c}}i{\'c}} {and}
  \bibinfo{person}{Markus Kr{\"o}tzsch}.} \bibinfo{year}{2014}\natexlab{}.
\newblock \showarticletitle{Wikidata: a free collaborative knowledgebase}.
\newblock \bibinfo{journal}{\emph{Commun. ACM}} \bibinfo{volume}{57},
  \bibinfo{number}{10} (\bibinfo{year}{2014}), \bibinfo{pages}{78--85}.
\newblock


\bibitem[Xie et~al\mbox{.}(2016)]%
        {xie2016representation}
\bibfield{author}{\bibinfo{person}{Ruobing Xie}, \bibinfo{person}{Zhiyuan Liu},
  \bibinfo{person}{Maosong Sun}, {et~al\mbox{.}}}
  \bibinfo{year}{2016}\natexlab{}.
\newblock \showarticletitle{Representation learning of knowledge graphs with
  hierarchical types.}. In \bibinfo{booktitle}{\emph{Proc. IJCAI 2016}},
  Vol.~\bibinfo{volume}{2016}. \bibinfo{pages}{2965--2971}.
\newblock


\bibitem[Xiong et~al\mbox{.}(2017)]%
        {xiong2017deeppath}
\bibfield{author}{\bibinfo{person}{Wenhan Xiong}, \bibinfo{person}{Thien
  Hoang}, {and} \bibinfo{person}{William~Yang Wang}.}
  \bibinfo{year}{2017}\natexlab{}.
\newblock \showarticletitle{Deeppath: A reinforcement learning method for
  knowledge graph reasoning}.
\newblock \bibinfo{journal}{\emph{arXiv preprint arXiv:1707.06690}}
  (\bibinfo{year}{2017}).
\newblock


\bibitem[Zhang et~al\mbox{.}(2020)]%
        {zhang2020learning}
\bibfield{author}{\bibinfo{person}{Zhanqiu Zhang}, \bibinfo{person}{Jianyu
  Cai}, \bibinfo{person}{Yongdong Zhang}, {and} \bibinfo{person}{Jie Wang}.}
  \bibinfo{year}{2020}\natexlab{}.
\newblock \showarticletitle{Learning hierarchy-aware knowledge graph embeddings
  for link prediction}. In \bibinfo{booktitle}{\emph{Proc. AAAI 2020}},
  Vol.~\bibinfo{volume}{34}. \bibinfo{pages}{3065--3072}.
\newblock


\bibitem[Zhong et~al\mbox{.}(2021)]%
        {zhong-etal-2021-factual}
\bibfield{author}{\bibinfo{person}{Zexuan Zhong}, \bibinfo{person}{Dan
  Friedman}, {and} \bibinfo{person}{Danqi Chen}.}
  \bibinfo{year}{2021}\natexlab{}.
\newblock \showarticletitle{Factual Probing Is [{MASK}]: Learning vs. Learning
  to Recall}. In \bibinfo{booktitle}{\emph{Proc. NAACL 2021}}.
  \bibinfo{publisher}{Association for Computational Linguistics},
  \bibinfo{address}{Online}, \bibinfo{pages}{5017--5033}.
\newblock
\urldef\tempurl%
\url{https://doi.org/10.18653/v1/2021.naacl-main.398}
\showDOI{\tempurl}


\bibitem[Zhu et~al\mbox{.}(2022)]%
        {zhu2022neural}
\bibfield{author}{\bibinfo{person}{Zhaocheng Zhu}, \bibinfo{person}{Mikhail
  Galkin}, \bibinfo{person}{Zuobai Zhang}, {and} \bibinfo{person}{Jian Tang}.}
  \bibinfo{year}{2022}\natexlab{}.
\newblock \showarticletitle{Neural-symbolic models for logical queries on
  knowledge graphs}. In \bibinfo{booktitle}{\emph{Proc. ICML 2022}}. PMLR,
  \bibinfo{pages}{27454--27478}.
\newblock


\end{thebibliography}

\appendix









\end{document}